\pdfoutput=1
\documentclass[11pt]{article}
\usepackage{overpic}
\usepackage{float}

\usepackage[a-1b]{pdfx}

\usepackage{framed}
\usepackage{lmodern}
\usepackage{mdwlist}
\usepackage{siunitx}
\usepackage{latexsym}
\usepackage{colortbl}
\usepackage{xcolor}
\usepackage{nicefrac}
\usepackage{booktabs}
\usepackage{fnpct}
\usepackage{amsfonts}
\usepackage[T1]{fontenc}
\usepackage{bold-extra}
\usepackage{amsmath}
\usepackage{amssymb}
\usepackage{bm}
\usepackage{graphicx}
\usepackage{mathtools}
\usepackage{microtype}
\usepackage{multirow}
\usepackage{multicol}
\usepackage{xpatch}
\usepackage{latexsym,comment}
\usepackage[normalem]{ulem}

\newcommand*{\missingreference}{{\Huge \colorbox{red}{?reference?}}}
\newcommand*{\missingcitation}{{\Huge \colorbox{red}{?citation?}}}

\makeatletter
\xpatchcmd{\@setref}{\bfseries}{\missingreference}{}{}
\def\@citex[#1]#2{\leavevmode
    \let\@citea\@empty
    \@cite{\@for\@citeb:=#2\do
        {\@citea\def\@citea{,\penalty\@m\ }%
            \edef\@citeb{\expandafter\@firstofone\@citeb\@empty}%
            \if@filesw\immediate\write\@auxout{\string\citation{\@citeb}}\fi
            \@ifundefined{b@\@citeb}{\hbox{\reset@font\missingcitation}%
                \G@refundefinedtrue
                \@latex@warning
                {Citation `\@citeb' on page \thepage \space undefined}}%
            {\@cite@ofmt{\csname b@\@citeb\endcsname}}}}{#1}}
\makeatother

\newcommand{\reasoning}{TTR}

\newcommand{\nli}[0]{NLI\xspace}

\newcommand{\mm}[0]{LLM\xspace}
\newcommand{\mcqa}{MCQA\xspace}

\newcommand{\mcq}{MCQ\xspace}

\newcommand{\gem}[1]{\mbox{\textsc{gem}}}
\newcommand{\abr}[1]{\textsc{#1}\xspace}

\newcommand{\D}[2]{\frac{\partial #1}{\partial #2}}

\newcommand{\hidetext}[1]{}
\newcommand{\ignore}[1]{}
\newif\ifcomment
\commentfalse

\ifcomment
    \newcommand{\pinaforecomment}[3]{\colorbox{#1}{\parbox{.8\linewidth}{#2: #3}}}

    \newcommand{\prtodo}[1]{\pinaforecomment{lightblue}{pr}{#1}}
    \newcommand{\prtodoi}[1]{\pinaforecomment{lightblue}{pr}{#1}}
\else
    \newcommand{\pinaforecomment}[3]{}
    \newcommand{\prtodo}[1]{}
    \newcommand{\prtodoi}[1]{}
\fi

\newcommand{\smallurl}[1]{ \begin{tiny}\url{#1}\end{tiny}}

\definecolor{lightblue}{HTML}{3cc7ea}
\definecolor{CUgold}{HTML}{CFB87C}
\definecolor{grey}{rgb}{0.95,0.95,0.95}
\definecolor{ceil}{rgb}{0.57, 0.63, 0.81}
\definecolor{UMDred}{HTML}{ed1c24}
\definecolor{UMDyellow}{HTML}{ffc20e}

\usepackage[]{acl2023}

\usepackage{xspace}
\usepackage{xcolor}
\usepackage[utf8]{inputenc}
\usepackage{pgfplots}
\usepackage{dsfont}
\DeclareUnicodeCharacter{2212}{−}
\usepgfplotslibrary{groupplots,dateplot}
\usetikzlibrary{patterns,shapes.arrows}
\pgfplotsset{compat=newest}

\usepackage{tikzscale}
\usepackage{relsize}
\usepackage{amsmath,amssymb}
\newcommand{\probP}{\text{I\kern-0.15em P}}

\usepackage{multirow, colortbl}

\usepackage{tabularx,booktabs}
\usepackage{makecell}

\usepackage[normalem]{ulem}
\useunder{\uline}{\ul}{}

\usepackage{graphicx}

\definecolor{ablation6}{HTML}{fcefed}
\definecolor{ablation_tie}{HTML}{fce3e1}

\definecolor{ablation5}{HTML}{fcd8d4}
\definecolor{ablation4}{HTML}{FBC3BC}
\definecolor{ablation3}{HTML}{F7A399}
\definecolor{ablation2}{HTML}{F38375}
\definecolor{ablation1}{HTML}{EF6351}

\usepackage[]{algpseudocode}
\usepackage[]{algorithm}
\usepackage{float}
\algtext*{EndFor}%
\algtext*{EndProcedure}%

\usepackage{booktabs}
\usepackage[normalem]{ulem}
\useunder{\uline}{\ul}{}

\usepackage{times}
\usepackage{latexsym}
\usepackage{adjustbox}

\usepackage[T1]{fontenc}
\usepackage{amsmath}
\usepackage{amssymb}
\usepackage{booktabs}
\usepackage{tikzscale}
\usepackage{amsmath}

\usepackage{multirow, colortbl}

\usepackage{tabularx,booktabs}
\usepackage{makecell}
\usepackage{multirow}
\usepackage{scalerel,xparse}
\usepackage{cleveref}
\usepackage{microtype}
\usepackage[most]{tcolorbox}

\usepackage{enumitem}
\crefformat{section}{\S#2#1#3}
\crefformat{subsection}{\S#2#1#3}
\crefformat{subsubsection}{\S#2#1#3}

\definecolor{bggray}{rgb}{0.95, 0.95, 0.95}
\usepackage[%
    framemethod=tikz,
    skipbelow=\topskip,
    skipabove=\topskip
]{mdframed}
\mdfsetup{%
    leftmargin=0pt,
    rightmargin=0pt,
    backgroundcolor=bggray,
    middlelinecolor=black,
    roundcorner=3
}

\definecolor{SkyBlue}{rgb}{0.53, 0.81, 0.92}

\newtcolorbox[
  list inside=prompt,
  auto counter,
  number within=section
]{prompt}[1][]{%
  enhanced,
  float*=t,                 % ← span both columns (top). Use float*=htbp if you want more placement options
  colbacktitle=black!60,
  fonttitle=\small,
  coltitle=white,
  fontupper=\footnotesize,
  boxsep=4pt,
  left=0pt, right=0pt, top=0pt, bottom=0pt,
  boxrule=1pt,
  width=\textwidth,          % full page width inside the spanning float
  enlarge left by=0mm,
  enlarge right by=0mm,
  listing only,
  listing options={
    basicstyle=\ttfamily\footnotesize,
    breaklines=true,
    breakatwhitespace=true,
    language=json
  },
  #1,
}

\newtcolorbox[
  list inside=trace,
  auto counter,
  number within=section
]{trace}[1][]{%
  enhanced,
  float*=t,
  colback=blue!5,             % light blue background
  colbacktitle=blue!60!black, % darker blue title bar
  colframe=blue!60!black,     % matching border
  fonttitle=\small,
  coltitle=white,
  fontupper=\footnotesize,
  boxsep=4pt,
  left=0pt, right=0pt, top=0pt, bottom=0pt,
  boxrule=1pt,
  width=\textwidth,
  enlarge left by=0mm,
  enlarge right by=0mm,
  listing only,
  listing options={
    basicstyle=\ttfamily\footnotesize,
    breaklines=true,
    breakatwhitespace=true,
    language=json
  },
  #1,
}

\definecolor{UMDred}{HTML}{ed1c24}

\definecolor{yellowcolor}{HTML}{ffc20e}
\definecolor{redcolor}{HTML}{e99999}
\definecolor{orangecolor}{HTML}{f6b26b}
\definecolor{yellowcolor}{HTML}{ffd966}
\definecolor{bluecolor}{HTML}{a0c5e8}
\definecolor{purplecolor}{HTML}{d9d2e9}

\title{Test-Time Reasoners Are Strategic Multiple-Choice Test-Takers}

\author{Nishant Balepur$^{1, 2}$  \hspace{0.5cm} \textbf{Atrey Desai}$^{1}$ \hspace{0.5cm}  \textbf{Rachel Rudinger}$^{1}$\\
  $^{1}$University of Maryland \hspace{0.6cm}
  $^{2}$New York University \hspace{0.6cm} \\
  \textcolor{black}{\texttt{\{nbalepur, rudinger\}@umd.edu}}, \hspace{0.5cm} \textcolor{black}{\texttt{adesai10@terpmail.umd.edu}} \\
}

\begin{document}
\maketitle

\begin{abstract} {
Large language models (LLMs) now give~reasoning before answering, excelling in~tasks like multiple-choice question answering~(\mcqa{}).
Yet, a concern is that \mm{}s do~not~solve \mcq{}s as intended, as work~finds \mm{}s sans reasoning succeed in \mcqa{} without using the question, i.e., choices-only.
Such partial-input success~is often linked to trivial shortcuts,~but reasoning traces could reveal if choices-only strategies are truly shallow.
To examine~these~strategies, we have reasoning \mm{}s solve \mcq{}s in~full and choices-only inputs; test-time reasoning often boosts accuracy in full and in choices-only, half the time.
While possibly due to shallow shortcuts, choices-only success is barely affected by the length of reasoning traces, and after finding traces pass faithfulness tests, we show they use less problematic strategies like inferring missing questions.
In all, we challenge claims that partial-input success is always a~flaw, so~we~propose how reasoning traces could separate~problematic data from less problematic reasoning.\footnote{Our code and data are available at: \url{https://github.com/nbalepur/mcqa-shortcuts}.}
}
\end{abstract}

\section{Introduction: When Cheaters Prosper} \label{section:intro}

Reasoning has become a central goal in Large Language Model (\mm{}) development \cite{xu2025towards}, with models now~designed to produce step-by-step traces before responding \cite{Zelikman2022STaRBRA}.
By scaling compute in test-time,~these models can aid complex, multi-step tasks like Deep Research \cite{shao-etal-2024-assisting}, software assistance \cite{wang2024openhands}, and scientific automation \cite{lu2024ai}.

Offline evaluations ensure models behave as intended~\cite{baker2025monitoring}. 
A common format is multiple-choice question answering \cite[\mcqa{}]{clark2020f}:~from a set of choices, pick the best answer to a question. 
While popular, work reveals non-reasoning \mm{}s beat random in \mcqa{} without the question \cite{Balepur2024ArtifactsOA}, i.e., choices-only, raising concerns that even stronger reasoning models might~amplify this behavior.

Conventional wisdom ascribes choices-only success to models exploiting shallow cues in~choices, suggesting they do so even with all inputs to bypass what \mcqa{} tests~\cite{chandak2025answer}.
However, past~\textsc{NLI}~work~shows partial-input success~does not entail ineptitude~in the original NLI task \cite{srikanth-rudinger-2022-partial}.
Analogously,~we~hypothesize \mm{}s may use non-shallow strategies in choices-only, drawing on skills \mcq{}s aim to test---e.g., discarding inaccurate choices---akin to how students with partial~knowledge guess on~MC exams \cite{lau2011guessing}. 
These are hard~to study in simple (e.g. bag-of-words) or encoder-only (e.g. BERT) models common in partial-input work \cite{poliak2018hypothesis}---unable to express them---but reasoning traces offer signals on how choices-only models could succeed.

Our short paper builds focused tests to target~this, running 12 \mm{}s in two settings: \textit{full input}---using questions and choices as input---and \textit{choices-only}---using just the choices.
In both, we study~the impact of test-time reasoning (\reasoning{}) on \mcqa{}~accuracy, aiming to learn: 1) if test-time reasoning~amplifies choices-only success; and 2) which types of strategies reasoning~traces use in choices-only settings.

In 36 \mm{}-benchmark combinations, \reasoning{} improves accuracy with full inputs in 27, but only 15 for choices-only.
Further, choices-only accuracy rises minimally with \reasoning{} scaling (\cref{subsection:scaling}), suggesting \mm{}s with and without \reasoning{} may use similar approaches to solving these items.
%\reasoning{} often improves accuracy in full \mcqa but just 15/36 cases in choices-only (\cref{subsection:accuracy}), and choices-only accuracy only slightly grows with longer reasoning (\cref{subsection:scaling}), so \reasoning{} does not largely amplify this potentially problematic behavior.
We then study if \mm{}s with \reasoning{} rely on shallow cues in choices-only; we ensure choices-only reasoning traces pass faithfulness checks (\cref{subsection:faithful}), then uncover they \textit{sometimes} employ superficial shortcuts that can bypass the original \mcqa{} task (e.g. \textit{I'll pick ``(A) 1.5'', as 1.5 is a ``messy'' number}), but \textit{more often} use less problematic strategies like inferring missing questions and naming properties of choices (\cref{subsection:qualitative})---still using skills the \mcq{} was meant to test or beyond.

Our mixed results challenge the typical view that partial-input success is always a flaw, with implications for researchers who both analyze model abilities and who create new benchmarks.
To draw this line, we show how merging choices-only reasoning analysis like ours with \mcq{} writing guidelines can help evaluators discern faulty data from less problematic reasoning, fixing the former (\cref{section:conclusion}).~We aim to spark discussion on ways to study partial inputs as \textsc{ai} models advance.
Our novel contributions~are:\\
\noindent \textbf{1)} The first experiments for the impact of test-time reasoning (\reasoning{}) on choices-only accuracy, showing gains from \reasoning{} are small and scale modestly, especially relative to standard full-input settings.\\
\noindent \textbf{2)} Analysis of \mm{} reasoning traces with choices-only to reveal \mm{}s can use a mix of simple shortcuts and less problematic choices-only strategies. \\
\noindent \textbf{3)} A new proposal for how evaluators can use our style of trace analysis to refine \mcq{} item quality.

\begin{figure*}
    \centering
    \includegraphics[width=\linewidth]{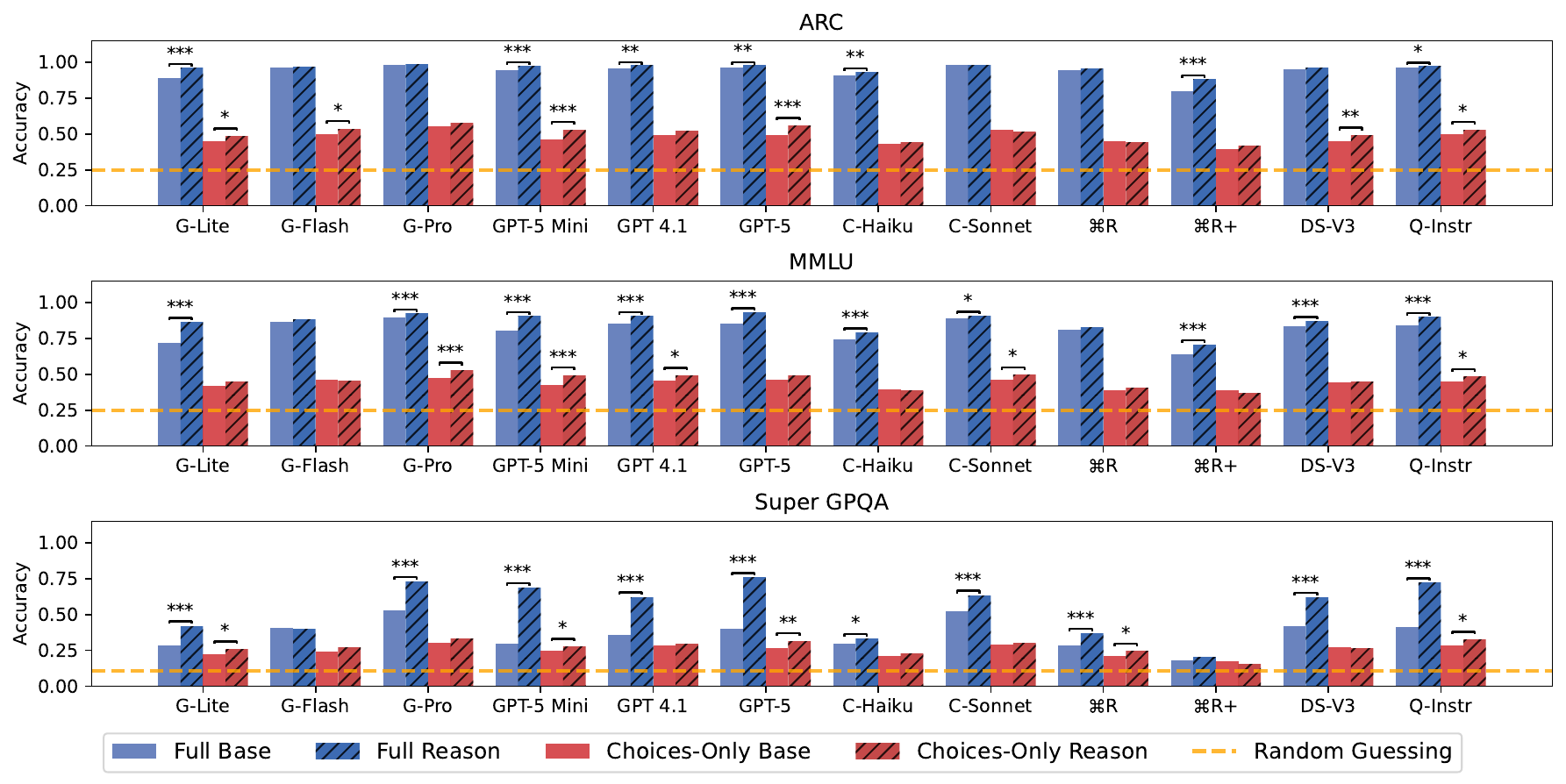}
    % \vspace{-5ex}
    \caption{\small \mm{} accuracy with/without reasoning in full/choices-only \mcqa. Command-R, Qwen, and DeepSeek do not support reasoning in the API, so we adjust it via prompt design (i.e., with/without chain-of-thought). * are significant differences \cite[paired t-test $p < 0.05$, $0.001$, $0.0001$]{student1908probable}. \reasoning~boosts Full accuracy in most cases and Choices-Only in 15/36 cases.  \label{fig:accuracy}}
    % \vspace{-1ex}
\end{figure*}

\section{Experiment Design} \label{section:setup}

We focus on \mcq{}s---a question $q$ and $n$ choices~$\mathcal{C}$ with one best answer $a \in \mathcal{C}$---and use two~settings to study reasoning \mm{}s' choices-only~accuracy:\\
\noindent \textbf{1) Full:} A typical \mcqa setup, where models use $q$ and $\mathcal{C}$ to select $a$ \cite{robinson2023leveraging}.\\
\noindent \textbf{2) Choices-Only:} Models see $\mathcal{C}$ and must infer~the right answer $a$ \cite{richardson2020does}.

We now offer datasets (\cref{subsection:dataset}), models (\cref{subsection:model}),~and prompts (\cref{subsection:prompt}) to isolate the effect of test-time reasoning \cite[\reasoning]{snell2024scaling} on these settings.

\subsection{Datasets} \label{subsection:dataset}

We evaluate on $1000$ randomly-sampled \mcq{}s in three popular benchmarks \cite{open-llm-leaderboard-v2}:\\
\textbf{1) ARC:} Testing grade-school scientific knowledge and commonsense reasoning \cite{clark2018think}.\\
\noindent \textbf{2) MMLU:} Testing knowledge in 57 college topics like math, history, or logic \cite{hendrycks2020measuring}.\\
\noindent \textbf{3) Super GPQA:} Testing graduate knowledge in 285 topics like engineering or law \cite{du2025supergpqa}.

\subsection{Models} \label{subsection:model}

We aim to examine \reasoning{} traces, so~we use $12$ \mm{}s with strong reasoning in $6$ families:
1) Gemini 2.5 \cite[Lite, Flash, Pro]{comanici2025gemini}; 
2) GPT (5 Mini, 4.1, 5); 
3) Claude (3.5 Haiku, 4 Sonnet); 
4) Cohere Command (R, R+); 
5) DeepSeek \cite[V3]{guo2025deepseek}; 
and 6) Qwen3 235B \cite[Instruct]{yang2025qwen3}.
We run all \mm{}s with LiteLLM.\footnote{https://www.litellm.ai/}

\subsection{Evaluation Configuration} \label{subsection:prompt}

We run \mm{}s (\cref{subsection:model}) via \textit{full} (access to $q$ and $\mathcal{C}$) and \textit{choices-only} (just access to $\mathcal{C}$) settings (\cref{section:setup}), scoring accuracy---if their predicted answer $\hat{a}$ matches the right answer $a \in \mathcal{C}$.
To study how \reasoning{} impacts accuracy, we design two prompts for each setting:\\
\textbf{1) Base:} the model selects $a$ directly without generating any reasoning (i.e. Full/Choices-Only Base).\\
\textbf{2) Reason:} the model gives step-by-step reasoning before picking $a$ (i.e. Full/Choices-Only Reason).

We use zero-shot prompts that explain each task, ask \mm{}s to use ``any strategy necessary'' to pick~$a$, and wrap $\hat{a}$ in ``<answer letter>'' (prompts in Appendix~\ref{appendix:experiment}). 
For \mm{}s that can support \reasoning{} in the API, we set reasoning effort to ``none'' for (1) and ``medium'' for (2).
Else, we prompt \mm{}s to give reasoning traces between ``<reasoning>'' for (2), i.e., chain-of-thought \cite{wei2022chain}.~We~use $1.0$ temperature and $81920$ max tokens.
Other hyperparemeters are default (details in Appendix~\ref{appendix:experiment}).

\section{Results} \label{section:results}

With our experiments (\cref{section:setup}), we now study how~test-time reasoning (\reasoning{}) impacts accuracy in full and choices-only \mcqa{} setups~(\cref{subsection:accuracy}).
We see if this scales with reasoning length (\cref{subsection:scaling}), then review reasoning traces to uncover strategies used in choices-only (\cref{subsection:qualitative}).
We conclude with a proposal for how our analysis can improve \mcqa{} benchmarks~(\cref{section:conclusion}).

\subsection{Reasoning \mm{}s Excel in Choices-Only} \label{subsection:accuracy}

\reasoning{} significantly boosts accuracy in \textit{full} \mcqa{} for 25/36 model-dataset settings (blue $\rightarrow$ dark blue, Fig~\ref{fig:accuracy}).
In \textit{choices-only}, such gains are weaker but present, with 15/36 improvements (red $\rightarrow$ dark red), so \reasoning{} weakly boosts choices-only accuracy.
All \mm{}s score well above random with just choices; GPT-5 hits $0.557$ on ARC.
Extending prior work on \mcqa{} partial inputs \cite{Balepur2024ArtifactsOA,chandak2025answer}, we show reasoning~\mm{}s~are still surprisingly accurate without using the question.

Accuracy gaps in full vs. choices-only vary by task difficulty.
ARC/MMLU's elementary/college-level \mcq{}s have wide accuracy gaps, suggesting in full settings, \mm{}s are not \textit{only} using choices-only cues to score highly.
In Super GPQA's graduate~\mcq{}s, some \mm{}s without~\reasoning{} (Base) score similarly in both settings: GPT-5 Mini, Haiku, and Command-R have small accuracy gaps, so~full accuracy~may stem from choices-only shortcuts.
But \reasoning{} (Reason) with these \mm{}s noticeably~widens gaps, so running \mm{}s+\reasoning{} on full and choices-only \mcqa{} settings can be one way to ensure~full task accuracy is not solely due to partial-input cues.

% We thus advise running \mm{}s with \reasoning{} on \mcqa{} datasets, as it quells concerns that full task accuracy stems from partial-input success.

\begin{figure}[t]
    \centering
    % \vspace{-2.75ex}
    \includegraphics[width=\linewidth]{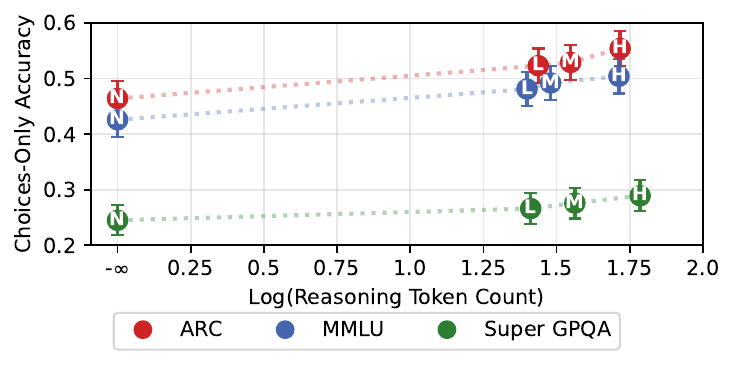}
    % \vspace{-5.5ex}
    \caption{\small GPT-5 Mini's choices-only accuracy and reasoning trace length across \textbf{L}ow, \textbf{M}edium, and \textbf{H}igh reasoning effort (95\% CIs). Longer reasoning slightly boosts model accuracy. \label{fig:reasoning}}
    % \vspace{-1.5ex}
\end{figure}

% may not Get Worse

\subsection{Longer Traces Slightly Aid Choices-Only} \label{subsection:scaling}

% The presence of partial-input success (\cref{subsection:accuracy}) leads to speculations that as models become more capable, they may become significantly better in partial-input settings \cite{yuan-etal-2024-llms}, but we show this is not always true.
% Within families, moving from the weakest GPT/Claude/Gemini \mm{}s to the strongest ones yields higher accuracy gains in full (avg $\Delta$ of 10.5\%)~than choices-only (avg $\Delta$ of 5.3\%, Fig \ref{fig:accuracy}).~Across time, \citet{Balepur2024ArtifactsOA} find LLaMA-2 70B \cite{touvron2023llama} at $0.67$ (full) and $0.38$ (choices-only) accuracy on MMLU, while two years later GPT-5 reaches $0.85$ (full) but only $0.46$ (choices-only).
% Currently, scaling \mm{} abilities improves accuracy on full inputs relatively more than partial inputs.

While we show \reasoning{} does not always boost choices-only accuracy (\cref{subsection:accuracy}), one concern is that we may have bound choices-only accuracy by fixing one setting for reasoning effort \cite[\cref{subsection:prompt}]{baker2025monitoring}.
To test this, we set GPT-5~Mini's reasoning effort to low/medium/high.\footnote{We replicate this in G-Flash/C-Sonnet (Appendix~\ref{appendix:extra_reasoning}).}
More effort~leads to longer traces (Fig~\ref{fig:reasoning}), but choices-only success only rises slightly.
This suggests \mm{}s may solve \mcq{}s~similarly across reasoning length, so traces may convey underlying strategies (\cref{subsection:qualitative}) versus accuracy gains.

\subsection{Reasoning Traces Are Informative} \label{subsection:faithful}

Before using \reasoning{} to study~choices-only strategies, we see whether they informatively support \mm{} answers.
Faithfulness is difficult to prove \cite{barez2025chain},~but~we~show traces pass three standard faithfulness checks \cite[Appendix~\ref{appendix:faithfulness}]{li-etal-2025-towards-better}:~1) after adding~\reasoning{},~\mm{}s maintain their answer far above chance, so it does not largely alter their behavior; 
2) GPT-5 predicts answers \mm{}s pick from their traces~with $>$90\% accuracy, so~traces are consistent with their selections; and
3) traces expose intentional cues we add to choices (e.g. duplicates), following \citet{turpin2024language}, meaning we do not detect traces as explicitly masking model behavior.
Given that our traces pass these standard faithfulness checks, we follow \citet{deng2025cot} and treat them as \textit{soft evidence}: informative signals to learn how choices-only success could \textit{potentially} arise.  

\begin{table*}[]
\small
\centering
\setlength{\tabcolsep}{3pt}
\renewcommand{\arraystretch}{1}
\begin{tabular}{@{}lll@{}}
\toprule
\textbf{Strategy} & \textbf{Description}              & \textbf{Example}                                                                                                              \\ \midrule
   FACT              & Recalling facts for choices       & \textit{Scanning the options, (A) "need energy to survive" leaps out as a universal truth}
                                         \\

ELIM              & Discarding inaccurate options      & \textit{Spiders eat insects, not grass, and rabbits eat plants, not mice.  This one is a mess}    \\
PATTERNS          & Naming properties in choices   & \textit{Looking at these, I immediately see patterns... three are non-renewable resources} \\
INFER Q           & Guessing the original question    & \textit{I'm confident the question is asking about the best way to ensure reliable results}              \\
SHALLOW           & Simple cues sans \mcq{} skills     & \textit{1.5 is the only value with a ``5'' in it. It's the ``messiest'' number}                                \\ \midrule
INCONS            & Trace does not support answer & \textit{...without knowing which, you can't make a definitive prediction.} $\rightarrow$ picks (D)                                                   \\ \bottomrule
\end{tabular}
% \vspace{-1ex}
\caption{\small Reasoning traces use $5$ choices-only strategies; all but SHALLOW use skills \mcq{}s normally test or beyond---less problematic than simple cues linked with partial inputs. We also see if traces inconsistently back answers, but this is rare (Fig~\ref{fig:qual}). \label{fig:examples}} 
% \vspace{-1ex}
\end{table*}

\subsection{Not All Strategies Are Problematic} \label{subsection:qualitative}

As \reasoning{} traces pass faithfulness checks (\cref{subsection:faithful}),~we now use them to see how \mm{}s may solve~\mcq{}s in choices-only.
We aim to learn: \textbf{Q1)}~what~strategies do traces convey in choices-only; and~\textbf{Q2)}~how strategies differ in choices-only successes/failures.

To answer these, we draw from human-computer interaction and use qualitative coding \cite{bingham2023data}. 
One author derives high-level themes from $180$~correct and incorrect choices-only traces from ARC, split evenly across $3$ \mm{}s with high choices-only success: G-Pro, C-Sonnet, and Q-Instr (Appendix~\ref{appendix:reasoning_trace} codes MMLU).
Each theme describes a high-level reasoning strategy used in the reasoning trace, found after two rounds of coding.
A second author validates themes by independently labeling 20 random examples, reaching $93$\% agreement.

% To answer these, we code \cite{bingham2023data} $180$~correct/incorrect choices-only traces in $3$ \mm{}s with high choices-only success---G-Pro, C-Sonnet, and Q-Instr---on ARC (Appendix~\ref{appendix:reasoning_trace} codes MMLU).  

\begin{figure}[!t]
    \centering
    \vspace{-2ex}
    \includegraphics[width=\linewidth]{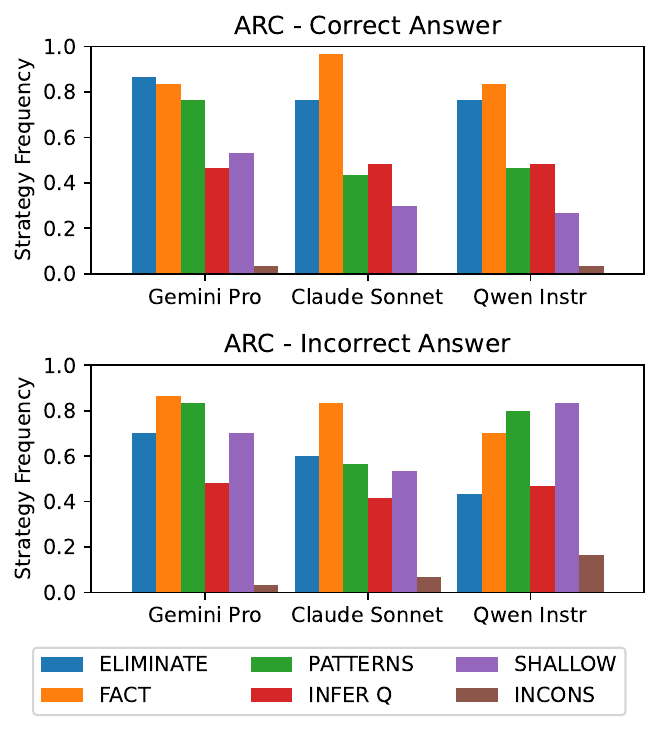}
    % \vspace{-5.25ex}
    \caption{\small Reasoning strategies when \mm{}s fail/succeed in choices-only. Traces use strategies beyond shallow~cues, so choices-only success is possible via~non-problematic means. \label{fig:qual}}
    % \vspace{-1.5ex}
\end{figure}

For Q1, we show $5$ strategies in Table~\ref{fig:examples} and~their prevalence in Fig~\ref{fig:qual}.\footnote{We repeat our analysis in the full \mcqa{} setting in Fig~\ref{fig:qual_parallel}.}
% For Q1, we find $5$ strategies:~\textbf{1) ELIM:} removing $c \in \mathcal{C}$ deemed wrong; \textbf{2) FACT:} recalling~facts for choices $\mathcal{C}$; \textbf{3) PATTERNS:} using dynamics~in~$\mathcal{C}$ (e.g. synonyms); \textbf{4) INFER Q:} trying to infer~the missing question; \textbf{5) SHALLOW:} trivial cues (e.g. (C) is often correct, ``apple'' is a common object); and \textbf{6) INCONS:} traces inconsistent with selections.\footnote{This rarely occurs, aligning with our faithfulness checks.}
Traces use SHALLOW cues, but also strategies using skills the \mcq{} was meant to test or beyond---recalling facts, naming choice properties, flagging inaccuracies, and guessing then answering the hidden question---so not all choices-only strategies bypass the original \mcq{}.
In fact, many reflect how students with partial~knowledge informatively guess on MC tests \cite{lau2011guessing}.

For Q2, we regress Fig~\ref{fig:qual}'s data to reveal~strategies predicting \mm{} choices-only success/failure (Appendix~\ref{appendix:reasoning_trace}); just SHALLOW and INCONS predict failure ($\alpha=0.05$), so correct \mm{} traces~use shallow cues less.
We then study~questions in INFER Q by assessing if they match the~original \mcq{} semantically.
In choices-only success, \mm{}s guess the original one 83\% of times in ARC (77\%, MMLU) but only 9\% for failure (13\%, MMLU).
We cannot rule out leakage, but no question exactly matches the original, and \mm{}s consistently try to infer~and solve MCQs in choices-only.
Thus, choices-only strategies can use skills \textit{beyond} what original \mcq{}s test, like abductive reasoning to find plausible explanations for the missing question \cite{zhao-etal-2023-abductive, balepur-etal-2025-reverse, balepur2025whose}.

In all, reasoning traces use a mix of shallow~and less problematic strategies.
We return in \cref{section:conclusion} to discuss how this tension can guide future evaluations.

\section{Related Work} \label{section:related_work}

\noindent \textbf{\mm{} Reasoning:} Reasoning is a long-standing~\textsc{ai} challenge \cite{mccarthy1960programs}, with rising~popularity and research in \mm{}s \cite{qiao-etal-2023-reasoning}.
Early~work~used prompting \cite{kojima2022large} to elicit~reasoning, while researchers now often train \mm{}s to always reason pre-response \cite{guo2025deepseek}.~Reasoning boosts accuracy~\cite{balepur-etal-2024-easy, sprague2025to} but can amplify risks including unfaithfulness~\cite{lyu2023faithful}, biases \cite{shaikh2022second}, and~user~deception \cite{williams2025on}; some are~\mcqa{}-specific, like miscalibration \cite{fu2025multiple} and~brittleness \cite{raman2025reasoning}.~We similarly~study~potential risks of reasoning in choices-only \mcqa{} settings.\\

\noindent \textbf{Partial-Input Models:} Partial-input models complete tasks with a subset of task inputs, like choices-only in \mcqa{} \cite{richardson2020does, balepur2024your}, hypothesis-only in \nli{} \cite{poliak2018hypothesis}, or text-only in vision \cite{goyal2017making}.
These models can find dataset artifacts \cite{gururangan2018annotation}, annotator biases \cite{geva-etal-2019-modeling}, and leakage \cite{gupta2024improving}.
Similar to us, \citet{Balepur2024ArtifactsOA} prompt~$4$ open-weight \mm{}s to show they succeed~in choices-only and speculate this involves inferring missing questions.
Instead, we move beyond prompt perturbations to reasoning analysis: in 12 reasoning \mm{}s, we show how test-time reasoning affects choices-only success and reveal a richer set of choices-only strategies (e.g. recalling facts, elimination, choice patterns), better disentangling shallow shortcuts from less problematic~reasoning.

\section{Conclusion: Rethinking Partial Inputs} \label{section:conclusion}

\mm{}s are surprisingly accurate in \mcqa{} with just choices.
Prior work often scolds partial-input~success as data/model flaws, but our reasoning study challenges this: some shallow strategies are clearly problematic, while others need skillful abilities the \mcq{} assesses or beyond, like inferring then solving missing questions.
Our work has implications for analysis researchers and benchmark creators.

\begin{figure}
    \centering
    \vspace{-1ex}
    \includegraphics[width=\linewidth]{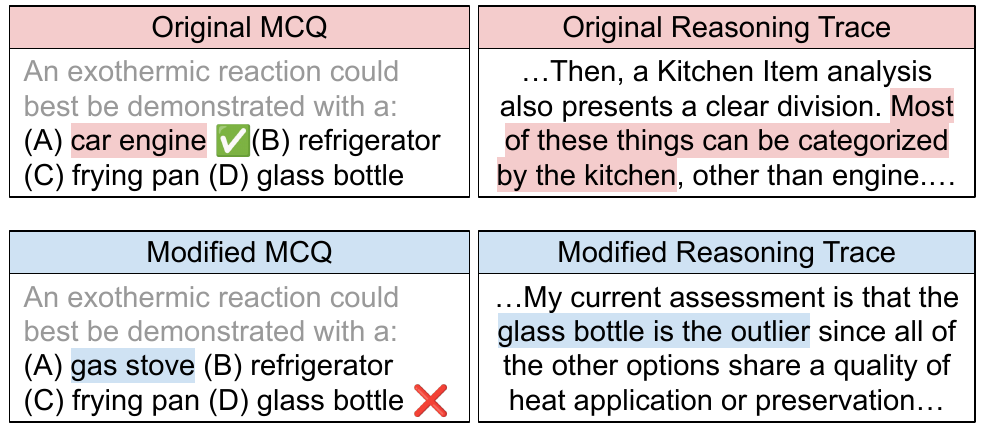}
    % \vspace{-4.5ex}
    \caption{\small \mcq{} solved by G-Pro in choices-only via shallowly picking the non-kitchen item. Changing the gold answer to a kitchen item (e.g. gas stove) removes partial-input success. \label{fig:case_study}}
    % \vspace{-1.5ex}
\end{figure}

\begin{itemize}
    \item \textbf{For analysis researchers:} Choices-only accuracy is an insufficient metric to determine whether a model or dataset is flawed, as such accuracy can be achieved via non-problematic means.
    We reveal how examinations of reasoning traces can disentangle whether benchmark items are problematic or whether the model expressed a more impressive capability (\cref{subsection:qualitative}), helping researchers rigorously analyze how often their datasets or models are flawed.

    \item \textbf{For benchmark creators:} Choices-only accuracy is still achievable using test-time reasoning models on modern benchmarks (Figure~\ref{fig:accuracy}), and such scores have amplified relative to prior work \citep{Balepur2024ArtifactsOA}. As this accuracy sometimes stems from shallow shortcuts (Figure~\ref{fig:qual}), we need more research on studying how to reduce these shortcuts in MCQA benchmarks, improving their validity.
\end{itemize}

To minimize shallow strategies' impact on \mcq{} evaluation, such researchers can merge trace analyses like ours with \mcq{} writing rules to fix benchmark items.
Figure~\ref{fig:case_study} illustrates a case study that instantiates such an intervention; an \mm{} initially exploits shallow cues in the \mcq{}---selecting the non-kitchen item.
This outlier violates \mcq{} writing rules \cite{haladyna2002review}, which advise making choices homogeneous in content; once fixed, choices-only success disappears.
While a~full evaluation design is outside a short paper's scope, we hope future work rethinks the role of partial-input studies in \mm{}s, with reasoning traces as signals~to split flawed data from less problematic reasoning.

\section{Limitations} \label{section:limitations}

While our setup reveals \mm{} reasoning typically improves accuracy in normal \mcqa{} and half the time in choices-only settings, we note \mm{}s are sensitive to prompts \cite{alzahrani2024benchmarks, shao2025spurious}.
Different prompts may change our results, so our reported choices-only accuracy forms a lower bound: it may rise with better prompts.
To address this issue, we ground our prompt design in best practices \cite{schulhoff2024prompt} and open tutorials for reasoning \mm{}s \cite{brown2025grpodemo} to ensure prompts are reasonable, and run significance tests for our experiments (Figure~\ref{fig:accuracy}).
Appendix~\ref{appendix:ablations} tests different prompt variations and Appendix~\ref{appendix:grpo} runs preliminary experiments with Supervised Fine-Tuning \cite{wei2021finetuned} and Group Relative Policy Optimization \cite{guo2025deepseek} to control for prompt sensitivity, consistent with Figure~\ref{fig:accuracy} results.

Further, while work has used \mm{} reasoning for safety monitoring \cite{baker2025monitoring}, we cannot ensure it is faithful \cite{agarwal2024faithfulness, van2024ai}.
However, we uncover that these reasoning traces pass faithfulness checks from prior work (\cref{subsection:faithful}): they contain few contradictions (\cref{subsection:qualitative}), support the model's decisions (Appendix~\ref{appendix:faithfulness}), and surface intentional perturbations (Appendix~\ref{appendix:faithfulness}).
This suggests models can offer a consistent explanation for their decisions, and like \citet{deng2025cot}, we reject the idea they have no useful signal at all.

\section{Ethical Considerations}

While the goal of \mm{} reasoning is to boost task accuracy, when applied to partial-input tasks---where the model must select the correct answer with ``any strategy necessary''---the system could rely on biases \cite{wu2025does}.
We believe this is unlikely in our \mcqa{} tasks testing factual knowledge, but we hope future work explores how such reasoning can amplify and mitigate biases \cite{chen2025reasoning}, fostering the development of safer \mm{} reasoning.

Generative AI (GenAI) was used in this project.
We used Cursor\footnote{https://cursor.com/agents} to design plots and refactor code, and GPT-5 to refine paper writing for brevity.
GPT-5 also converted a PDF \cite{deng2025cot} into the correct bibTeX format, which was meticulously reviewed for accuracy.
We never use GenAI for qualitatively coding data or writing text from scratch in this paper.
We take complete responsibility for any GenAI errors.
By discussing GenAI usage here, we aim to encourage other researchers to do the same.

\section*{Acknowledgments}

We would like to thank the \abr{clip} lab at the University of Maryland and our external collaborators for their help.
In particular, we thank Navita Goyal, Shashwat Goel, Nikhil Chandak, Shi Feng, and Jordan-Boyd Graber for reviews and discussions on earlier versions of this paper.
This material is based upon work supported by the National Science Foundation under \abr{iis}-2339746 (Rudinger) and \abr{dge}-2236417 (Balepur).
Any opinions, findings, and conclusions or recommendations expressed in this material are those of the author(s) and do not necessarily reflect the views of the National Science Foundation.
Access to Cohere's models~was made possible with a Cohere for AI Research Grant.

\bibliography{custom}
\bibliographystyle{acl_natbib}

\clearpage

\appendix
\section{Appendix} \label{section:appendix}

\subsection{Dataset Details} \label{appendix:dataset}

We sample $1000$ random examples from the test set of ARC \cite{clark2018think}, MMLU \cite{hendrycks2020measuring}, and Super GPQA \cite{du2025supergpqa}.
All datasets are publicly available, so our experiments are within their intended use.
We did not collect any datasets, so we did not check for PII.
To our knowledge, all questions are in English.

\subsection{Experiment Details} \label{appendix:experiment}

We access all \mm{}s with LiteLLM\footnote{https://docs.litellm.ai/} through their native APIs (e.g. the OpenAI API for GPT-5).
For Deepseek and Qwen, we use TogetherAI\footnote{https://www.together.ai/}.
We allocate 72 CPU hours for each experiment.
All results are reported from a single run.
Our prompts for thinking \mm{}s in full and choices-only settings are in Prompt~\ref{prompt:full} and Prompt~\ref{prompt:partial}, while our prompts for non-thinking \mm{}s in full and choices-only settings are in Prompt~\ref{prompt:full_cot} and Prompt~\ref{prompt:partial_cot}.
``input\_text'' is infilled with $q$ and $\mathcal{C}$ accordingly.
The endpoints for the models we used are:
\begin{itemize}[noitemsep]
    \item gemini/gemini-2.5-flash-lite
    \item gemini/gemini-2.5-flash
    \item gemini/gemini-2.5-pro
    \item openai/gpt-5-2025-08-07
    \item openai/gpt-5-mini-2025-08-07
    \item openai/gpt-5-nano-2025-08-07
    \item openai/gpt-4.1-2025-04-14
    \item anthropic/claude-sonnet-4-20250514
    \item anthropic/claude-3-5-haiku-20241022
    \item together\_ai/deepseek-ai/DeepSeek-V3
    \item together\_ai/deepseek-ai/DeepSeek-R1
    \item together\_ai/Qwen/Qwen3-235B-A22B-Instruct-2507-tput
    \item together\_ai/Qwen/Qwen3-235B-A22B-Thinking-2507
\end{itemize}

\subsection{Extra Open-Source Reasoning Models} \label{appendix:extra_models}

We also ran experiments with two popular open-source thinking models: DeepSeek r1 \cite{guo2025deepseek} and Qwen-3 Thinking \cite{yang2025qwen3}.
However, we found it difficult to disable thinking in these models, even after appending ``<think>Okay I have finished thinking.</think>'' to the prompt, as suggested by \citet{ma2025reasoning}.
We present results for these models in Figure~\ref{fig:accuracy_extra}, but note that ``Base'' reasoning traces still convey extensive reasoning, as they could not be completely disabled.

\subsection{Faithfulness Experiments} \label{appendix:faithfulness}

\begin{figure*}
    \centering
    \includegraphics[width=\linewidth]{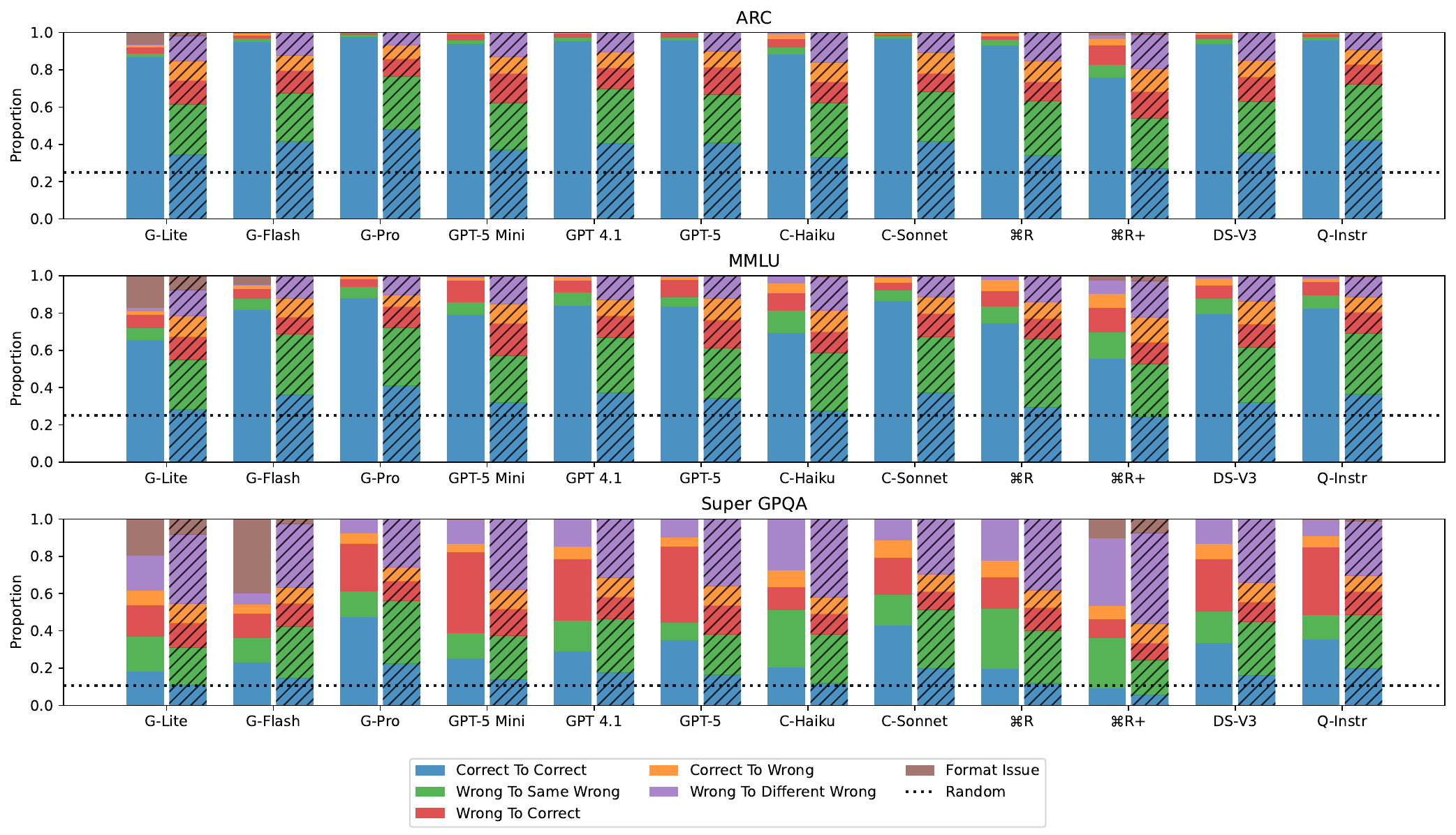}
    \caption{Different cases where \mm{}s change/maintain their answer selection after adding \reasoning{} in full (unhatched) and choices-only (hatched) settings. \mm{}s are consistent (blue+green bars) well-above random (black dotted line). \label{fig:swap}}
\end{figure*}

This section designs experiments to test the faithfulness of \mm{} reasoning traces.
We first examine how often \mm{}s swap their decisions after adding \reasoning{}.
If \mm{}s swap their decisions as much as a random baseline---selecting choices uniformly---\reasoning{} would not faithfully explain the model's decisions without \reasoning{}.
However, as shown in Figure~\ref{fig:swap}, most \mm{}s maintain their decisions after adding \reasoning{}---significantly above random---so we cannot claim \reasoning{} is unfaithful.

\begin{figure*}
    \centering
    \includegraphics[width=\linewidth]{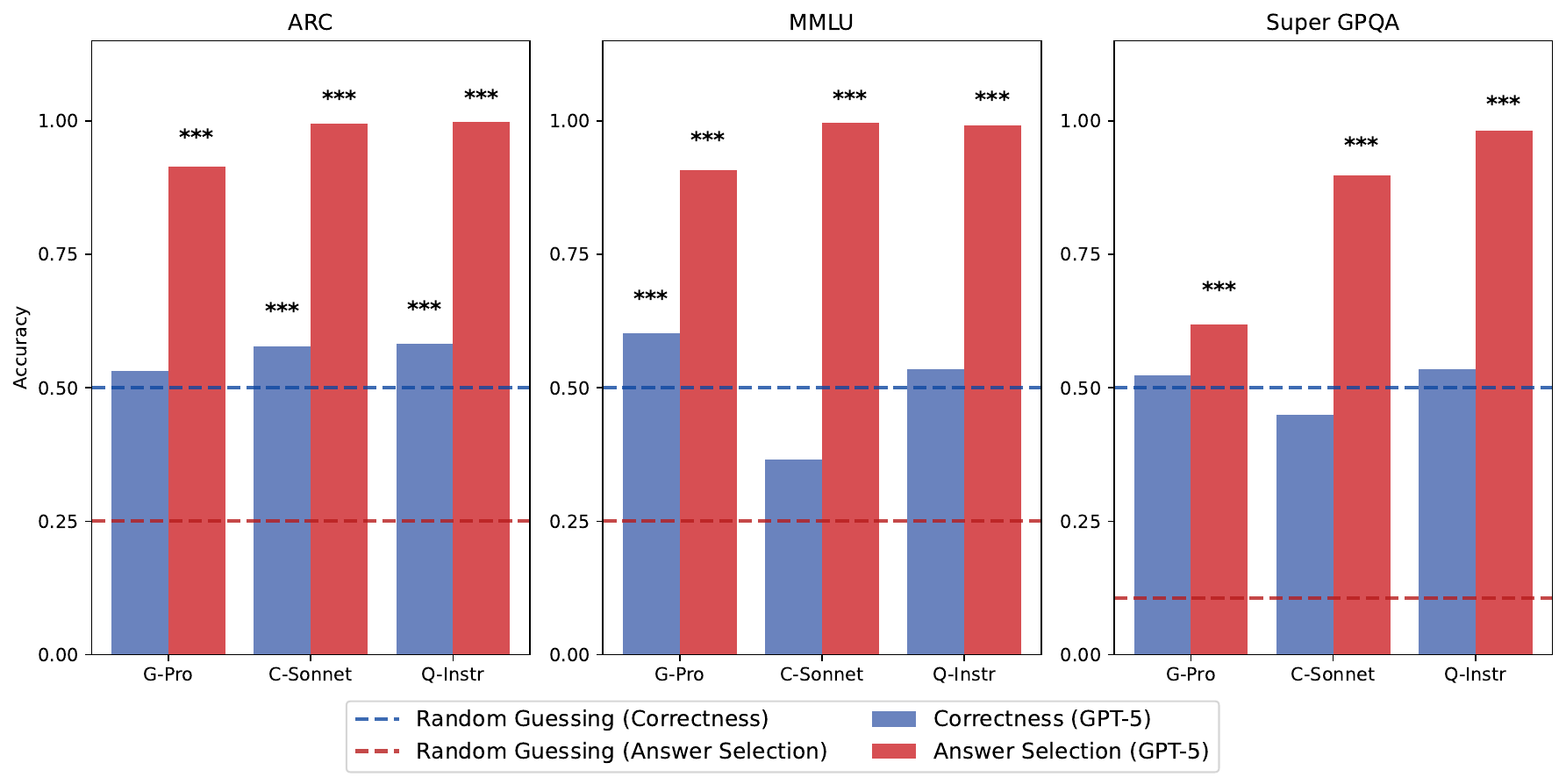}
    \caption{Accuracy of GPT-5 for predicting the model's selected answer and if the model was correct based on its reasoning trace in the choices-only setting. The model always surpasses random in the former, but cannot exceed random in the latter. Thus, reasoning traces are consistent with the selected answer, but do not convey obvious cues GPT-5 can pick up on to discern if the model arrived at the correct or incorrect answer in choices-only. \label{fig:faithful_pred}}
\end{figure*}

Next, a common issue with unfaithful reasoning traces is that they do not support the final answer \cite{barez2025chain}, so we test if this occurs in choices-only.
We analyze this on a subset of traces in \cref{subsection:qualitative}, but to assess this at scale, we see if GPT-5 can predict the selected answer choice of models just from their traces.
We test this with G-Pro, C-Sonnet, and Qwen-Instr's reasoning traces---the models we analyze in \cref{subsection:qualitative}.
If GPT-5 does no better than random, \reasoning{} would not faithfully explain the model's decisions.
But again, Figure~\ref{fig:faithful_pred} shows GPT-5's accuracy exceeds $0.90$, so we cannot claim \reasoning{} is unfaithful.
We also test if GPT-5 can predict if reasoning traces lead to correct/incorrect answers, which could let us find features indicative of accurate/inaccurate reasoning, but GPT-5 was unable to flag these differences (Figure~\ref{fig:faithful_pred}).

Finally, we follow \citet{turpin2024language} and modify $42$ \mcq{}s from ARC that G-Pro, C-Sonnet, and Qwen-Instr answer correctly with one of four perturbations to the choices: 1) duplicating the correct answer choice (Trace~\ref{trace:duplicate}); 2) adding a synonym of the correct answer choice (Trace~\ref{trace:synonym}); 3) adding a nonsensical choice in place of the selection (Trace~\ref{trace:random}); and 4) making the correct answer choice a factually incorrect statement (Trace~\ref{trace:inaccurate}).
On these questions, if the model does not change its decision and does not surface these perturbations in its reasoning trace, we can claim the reasoning trace is unfaithful.
But once again, G-Pro, C-Sonnet, and Qwen-Instr \textbf{always} either switch their answer or articulate the perturbation, so we cannot claim \reasoning{} is unfaithful.

\mm{}s pass all of our faithfulness checks, suggesting they can form ``soft evidence'' for studying strategies in choices-only settings.

\subsection{Extended Reasoning Length Analysis} \label{appendix:extra_reasoning}

\begin{figure*}
    \centering
    \includegraphics[width=\linewidth]{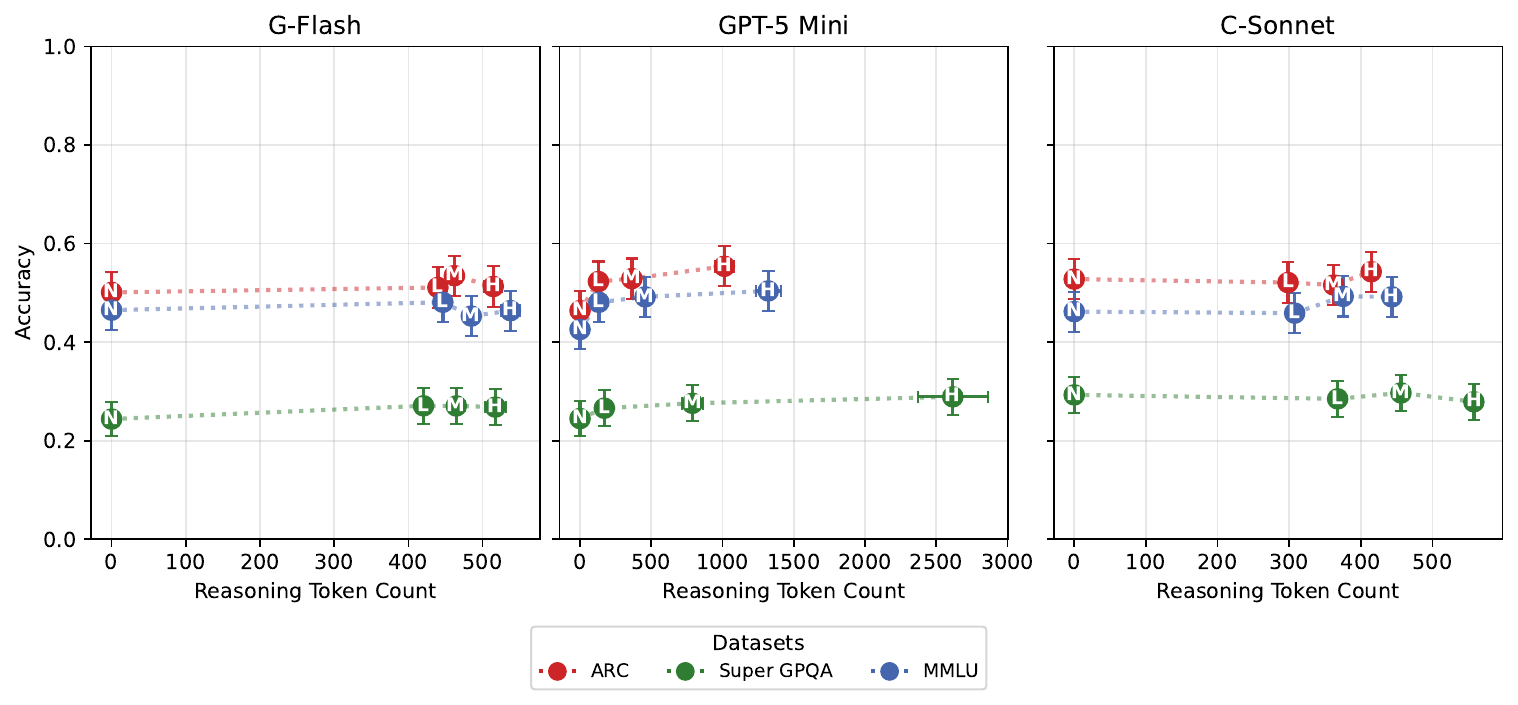}
    \caption{\small Full plot for reasoning length versus \textbf{choices-only} accuracy on GPT-5 Mini, Claude-4 Sonnet, and Gemini-2.5 Flash. The trend is consistent with Figure~\ref{fig:reasoning}: significantly increasing reasoning length does not significantly boost choices-only accuracy. \label{fig:reasoning_extra-partial}}
\end{figure*}

\begin{figure*}
    \centering
    \includegraphics[width=\linewidth]{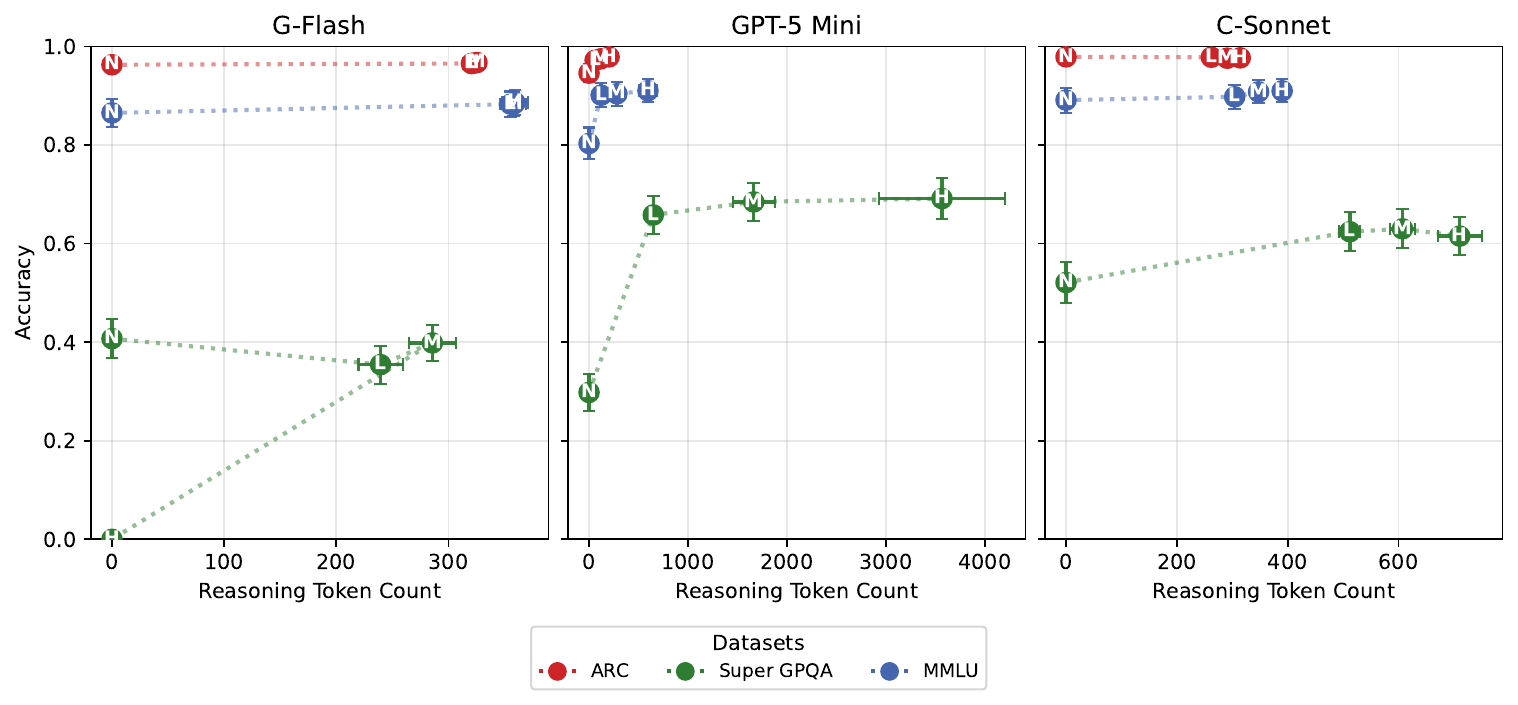}
    \caption{\small Full plot for reasoning length versus \textbf{full} \mcqa{} accuracy on GPT-5 Mini, Claude-4 Sonnet, and Gemini-2.5 Flash. Interestingly, significantly scaling reasoning length in normal \mcqa{} also does not significantly improve accuracy. \label{fig:reasoning_extra-full}}
\end{figure*}

We provide full versions of the plot in Figure~\ref{fig:reasoning} for GPT-5 Mini with two more \mm{}s (Gemini-2.5 Flash, Claude-4 Sonnet) in choices-only (Figure~\ref{fig:reasoning_extra-partial}).
While reasoning trace length does not increase as much in Gemini and Claude when setting ``reasoning effort'' to low, medium, and high, the trend remains: reasoning trace length significantly increases but choices-only accuracy does not significantly increases.
Interestingly, the same trend is shown in the full setting (Figure~\ref{fig:reasoning_extra-full}), so scaling reasoning length may not generally help \mm{}s answer \mcq{}s \cite{hassid2025don}.

\subsection{Extended Reasoning Trace Analysis} \label{appendix:reasoning_trace}

\begin{figure*}
    \centering
    \includegraphics[width=\linewidth]{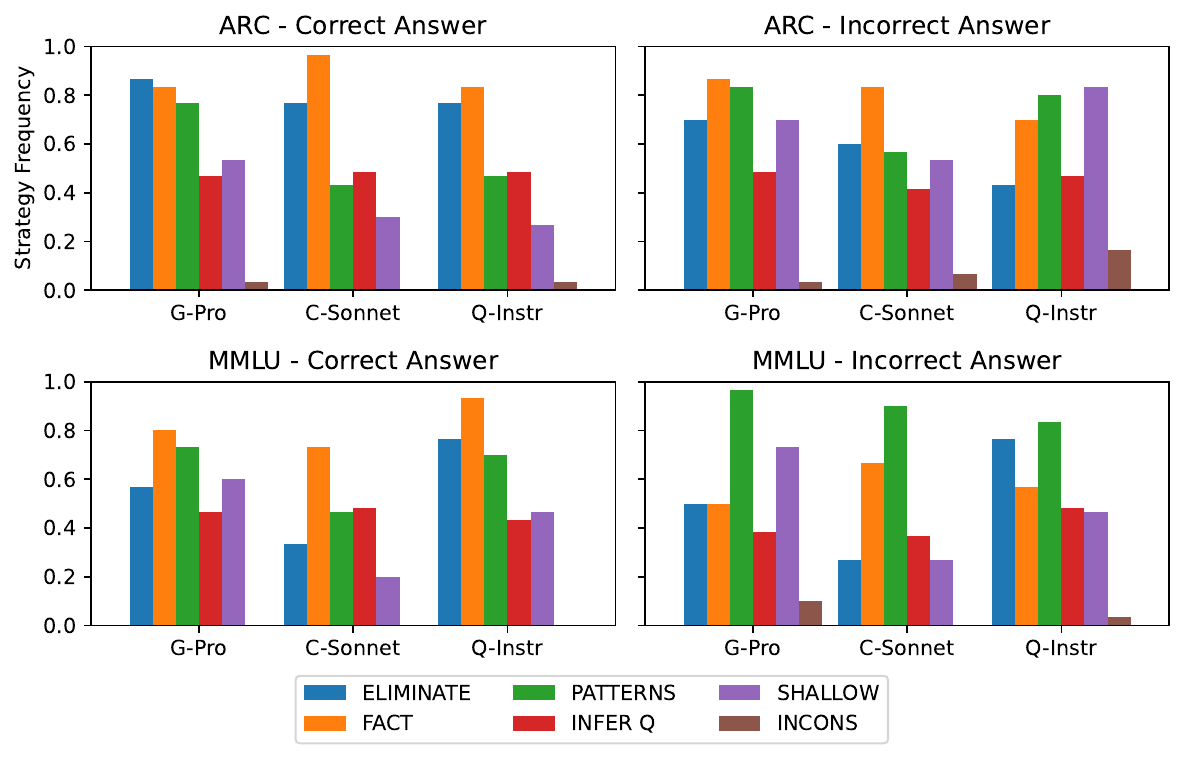}
    \caption{Full qualitative evaluation for ARC and MMLU. Trends are consistent with Figure~\ref{fig:qual}: \mm{} reasoning traces often use strategies beyond just simple cues in choices-only settings. \label{fig:qual_extra}}
\end{figure*}

While we discuss ARC in \cref{subsection:qualitative}, Figure~\ref{fig:qual_extra} repeats our qualitative coding procedure on MMLU, where trends are consistent: \mm{}s often use strategies beyond just shallow cues in partial-input settings.

For our regression, we use statsmodels\footnote{https://www.statsmodels.org/stable/index.html} to fit a logistic regression of choices-only success based on binary indicators for if a strategy is prseent ina  reasoning trace.
We control for dataset and model via categorical fixed effects with constants (i.e., add C(dataset) and C(model) for each model).
We run regressions separately for ARC and MMLU, with results in Table~\ref{appendix:table_arc_regression} and Table~\ref{appendix:table_mmlu_regression}, respectively.
The regression on MMLU shares similarities with ARC: no strategy is predictive of choices-only success/failure, so \mm{} traces consistently use non-problematic strategies regardless of whether they are arriving at the correct or incorrect answer. 

\begin{figure*}
    \centering
    \includegraphics[width=\linewidth]{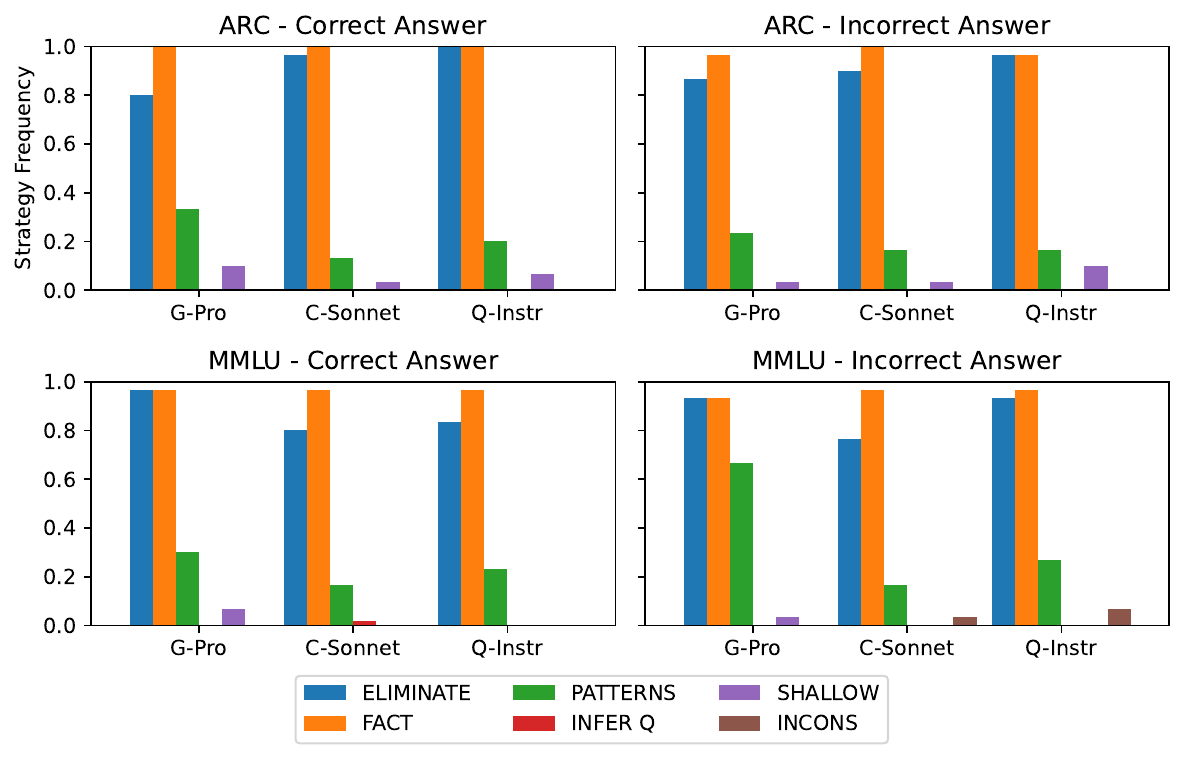}
    \caption{Qualitative evaluation for ARC and MMLU reasoning traces when choices-only succeeds/fails, but in the full setting. \mm{} full reasoning traces still often use strategies also found in choices-only. \label{fig:qual_parallel}}
\end{figure*}

Finally, we run another qualitative analysis, where we take the same \mcq{}s used in our choices-only trace analysis, but study \mm{}s' full reasoning traces (Figure~\ref{fig:qual_parallel}).
Interestingly, \mm{} these traces still use some of the strategies shown in choices-only---like recalling facts, eliminating options, and naming patterns in choices.
As expected, the \mm{} never tries to guess the question, as it already has access to it.
We also see signs of shallow shortcuts, but note that these normally occur after the \mm{} already solved the \mcq{}, forming a sanity check; for example, in a question about turtle speed, the \mm{} noted that some speeds were ``too fast'' to be plausible, while for other questions, they noted the answer was a ``textbook example''.
To curb these shortcuts, perhaps \mcq{}s could be rewritten as in \cref{section:conclusion}---forcing the \mm{} to solve the \mcq{} without commonsense knowledge (e.g., a math \mcq{} where a turtle ends up being very quick as the answer).

\subsection{SFT vs GRPO Comparison} \label{appendix:grpo}

\begin{figure*}
    \centering
    \includegraphics[width=\linewidth]{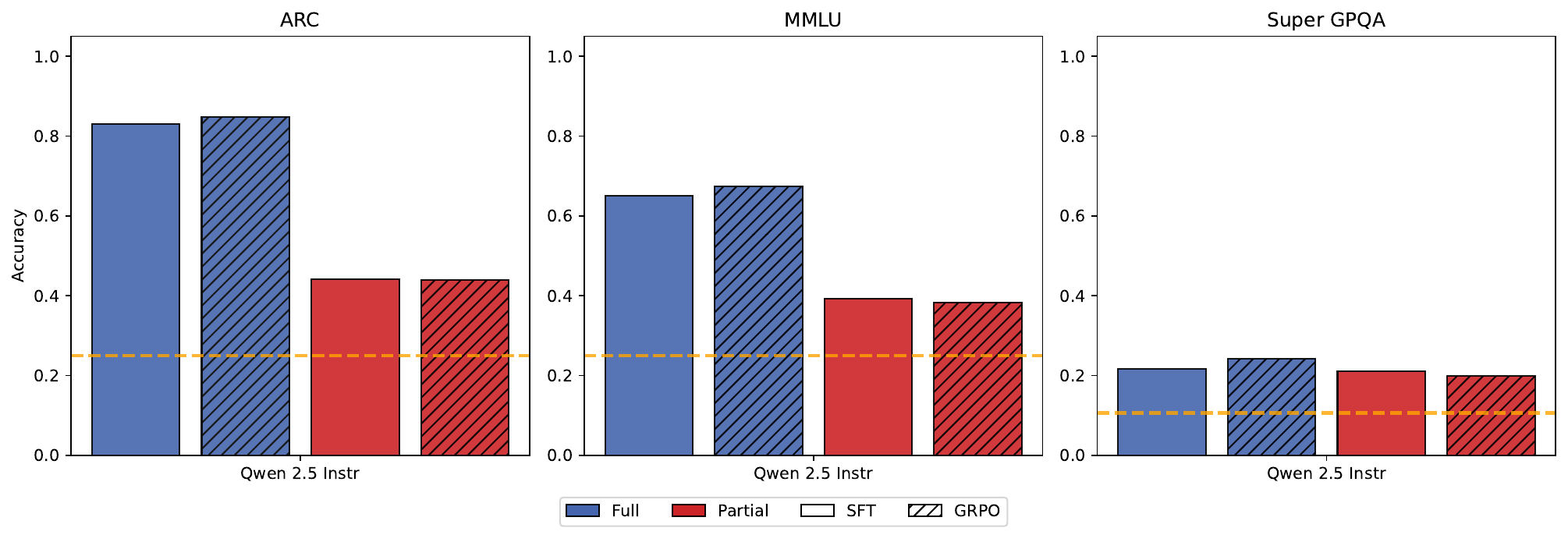}
    \caption{Full and choices-only accuracy of Supervised Fine-Tuning \cite[SFT]{longpre2023flan} versus Group Relative Policy Optimization \cite[GRPO]{guo2025deepseek} with Qwen-2.5 Instruct (3B). Mirroring Figure~\ref{fig:accuracy}, both strategies exceed random, and adding reasoning via GRPO improves full accuracy more than choices-only accuracy.  \label{fig:grpo}}
\end{figure*}

To study how different training strategies impact full and choices-only accuracy, we fine-tune Qwen-2.5 Instruct \cite[3B]{team2024qwen2} with two strategies: Supervised Fine-Tuning \cite[SFT]{longpre2023flan} and Group Relative Policy Optimization \cite[GRPO]{guo2025deepseek}.
SFT optimizes the \mm{} to directly predict the correct answer, while GRPO rewards the \mm{} for producing reasoning traces that lead to the correct answer.\footnote{We model our \mcqa{} rewards following the tutorial in: https://huggingface.co/learn/cookbook/fine\_tuning\_llm\_grpo\_trl}
In Figure~\ref{fig:grpo}, we show both strategies exceed random accuracy on all datasets, but GRPO does not largely exceed SFT in choices-only settings.
This further confirms our claims in \cref{subsection:accuracy} and \cref{subsection:scaling}---scaling \mm{} reasoning may not always amplify partial-input success.

\subsection{Prompt Ablations} \label{appendix:ablations}

We run ablations of different prompt designs to assess their impact on our experiments.
We first compare our analysis to a version of our \mcqa{} task where one of the options is ``I don't know'' (IDK).
Perhaps if IDK was added to the options, the model would abstain in choices-only, believing the task to be impossible.
We tested this in Tables~\ref{appendix:arc_idk} and \ref{appendix:mmlu_idk}; after adding ``I don’t know'' as choice (E), accuracy consistently drops by a few points, similar to work showing accuracy drops after adding ``None of the above'' \citep{elhady2025wicked}. However, choices-only accuracy still exceeds random (0.25), suggesting LLMs have enough confidence on the correct answer from the choices to avoid ``IDK''.

Next, we compare the prompt we design (\cref{subsection:prompt})---instructing models to use ``any strategy necessary'' to derive the answer---versus a more standard prompt from the InspectAI evaluation library \citep{inspect_ai_framework}.
Tables~\ref{appendix:arc_prompts} and \ref{appendix:mmlu_prompts} show  in the choices-only setting with reasoning, changing the prompt to InspectAI's prompt slightly drops accuracy for Gemini and GPT-5; our preliminary analysis finds this mainly stems from models abstaining, which our instruction of ``use any strategy necessary'' attempts to curb.
Still, all scores are much higher than random (0.25), so models still score highly in choices-only and motivating our analysis of LLM reasoning traces.

\subsection{Qualitative Examples} \label{appendix:qualitative}

This section provides examples of \mm{} reasoning traces in choices-only settings on ARC.
We show cases where \mm{}s can infer the original question (Trace~\ref{trace:itq}), pick up on differences in specificity indicating a dataset flaw (Trace~\ref{trace:flaw}), and reason over dynamics of choices (Trace~\ref{trace:patterns}).

% \section{Generative AI Usage Statement} \label{appendix:genai}

% Generative AI (GenAI) was used in this project.
% We used Cursor\footnote{https://cursor.com/agents} to design plots and GPT-5 to refine paper writing for brevity.
% We never use GenAI for writing experimentation code, qualitatively coding data, or writing text from scratch in this paper.
% We take full responsibility for any issues from GenAI errors.
% By discussing GenAI usage here, we aim to encourage other researchers to do the same.

\begin{figure*}
    \centering
    \includegraphics[width=\linewidth]{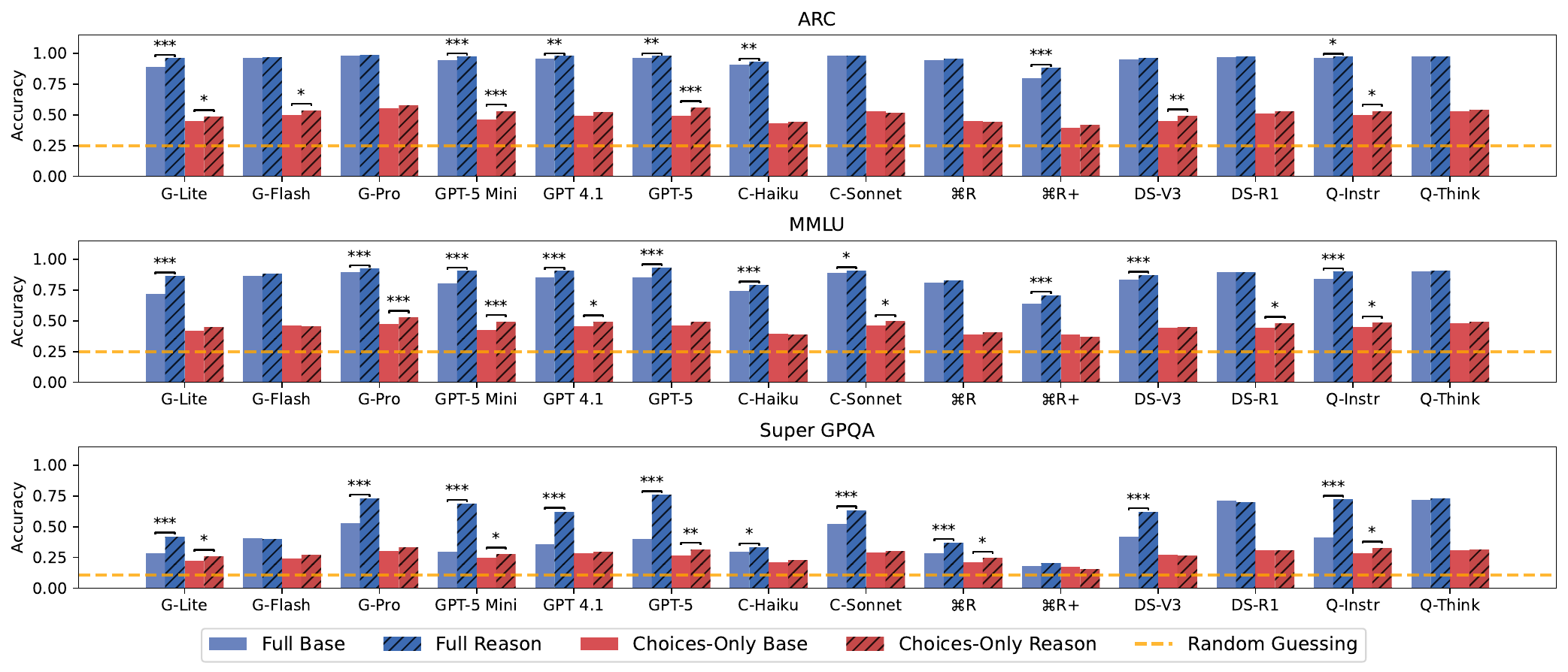}
    \caption{\small \mm{} accuracy with/without reasoning in full/choices-only \mcqa settings with the addition of DeepSeek-r1 and Qwen-3 Thinking. We note that we were unable to disable ``reasoning'' completely in the Base settings. \label{fig:accuracy_extra}}
\end{figure*}
\begin{figure*}
    \centering
    \includegraphics[width=\linewidth]{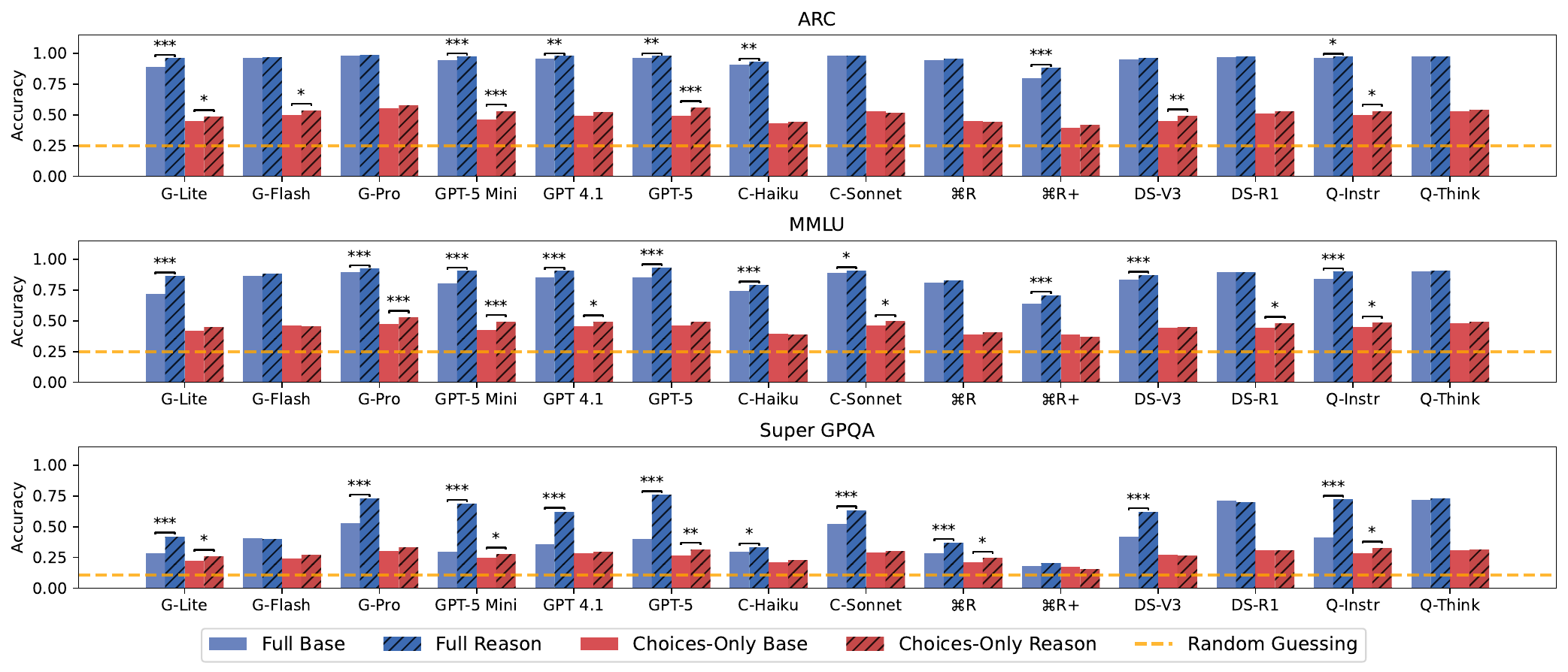}
    \caption{\small \mm{} accuracy with/without reasoning in full/choices-only \mcqa settings with the addition of DeepSeek-r1 and Qwen-3 Thinking. We note that we were unable to disable ``reasoning'' completely in the Base settings. \label{fig:accuracy_extra}}
\end{figure*}



\begin{table*}[htbp]
\centering
\begin{tabular}{lrrrrrr}
\toprule
 & Coef. & Std. Err. & z & P$>$|z| & 0.025 & 0.975 \\
\midrule
Intercept          &  0.046 & 0.113 &  0.404 & 0.686 & -0.175 &  0.266 \\
C(model)[T.Qwen]   &  0.036 & 0.139 &  0.257 & 0.797 & -0.237 &  0.309 \\
C(model)[T.gemini] &  0.031 & 0.140 &  0.225 & 0.822 & -0.242 &  0.305 \\
ELIMINATE          &  0.258 & 0.200 &  1.289 & 0.197 & -0.134 &  0.649 \\
FACT               &  0.026 & 0.182 &  0.141 & 0.888 & -0.332 &  0.383 \\
INCONS             & -1.460 & 0.795 & -1.835 & 0.067 & -3.019 &  0.099 \\
INFER\_Q           & -0.021 & 0.175 & -0.118 & 0.906 & -0.363 &  0.322 \\
PATTERNS           & -0.348 & 0.205 & -1.695 & 0.090 & -0.750 &  0.054 \\
SHALLOW            & -0.701 & 0.231 & -3.035 & 0.002 & -1.153 & -0.248 \\
\bottomrule
\end{tabular}
\caption{\small Logistic regression to study which reasoning trace strategies predict choices-only/failure success on ARC.}
\label{appendix:table_arc_regression}
\end{table*}

\begin{table*}[htbp]
\centering
\begin{tabular}{lrrrrrr}
\toprule
 & Coef. & Std. Err. & z & P$>$|z| & 0.025 & 0.975 \\
\midrule
Intercept          & -0.014 & 0.108 & -0.133 & 0.894 & -0.226 &  0.197 \\
C(model)[T.Qwen]   &  0.005 & 0.140 &  0.033 & 0.974 & -0.270 &  0.279 \\
C(model)[T.gemini] &  0.032 & 0.140 &  0.227 & 0.820 & -0.243 &  0.306 \\
ELIMINATE          &  0.085 & 0.220 &  0.385 & 0.700 & -0.347 &  0.517 \\
FACT               &  0.356 & 0.197 &  1.804 & 0.071 & -0.031 &  0.742 \\
INCONS             & -22.754 & 43399.281 & -0.001 & 1.000 & -8.51e+04 & 8.50e+04 \\
INFER\_Q           &  0.117 & 0.178 &  0.658 & 0.511 & -0.232 &  0.466 \\
PATTERNS           & -0.350 & 0.190 & -1.844 & 0.065 & -0.723 &  0.022 \\
SHALLOW            & -0.149 & 0.236 & -0.632 & 0.528 & -0.613 &  0.314 \\
\bottomrule
\end{tabular}
\caption{\small Logistic regression to study how the presence of reasoning strategies predict choices-only success on MMLU.}
\label{appendix:table_mmlu_regression}
\end{table*}

\begin{table*}[h]
\centering
\setlength{\tabcolsep}{4pt}
\small
\begin{tabular}{lcccc}
\toprule
Model & Choices-Only Base & Choices-Only Base (+ IDK) & Choices-Only Reason & Choices-Only Reason (+ IDK) \\
\midrule
G-Lite        & 0.4530 & 0.4597 & 0.4840 & 0.4655 \\
G-Flash       & 0.5010 & 0.4880 & 0.5350 & 0.5075 \\
GPT-5 Mini    & 0.4640 & 0.4273 & 0.5290 & 0.4848 \\
GPT 4.1       & 0.4910 & 0.4945 & 0.5180 & 0.5080 \\
Claude Haiku  & 0.4330 & 0.4523 & 0.4440 & 0.4525 \\
Command R     & 0.4530 & 0.4075 & 0.4440 & 0.4500 \\
\bottomrule
\end{tabular}
\caption{Evaluation of choices-only accuracy on ARC when adding ``I don't know'' (IDK) as one of the choices, similar to \citet{wang2025llms}. Accuracy still exceeds random guessing. \label{appendix:arc_idk}}
\end{table*}

\begin{table*}[h]
\centering
\small
\setlength{\tabcolsep}{4pt}
\begin{tabular}{lcccc}
\toprule
Model & Choices-Only Base & Choices-Only Base (+ IDK) & Choices-Only Reason & Choices-Only Reason (+ IDK) \\
\midrule
G-Lite        & 0.4170 & 0.4020 & 0.4440 & 0.4353 \\
G-Flash       & 0.4650 & 0.4515 & 0.4530 & 0.4698 \\
GPT-5 Mini    & 0.4258 & 0.3907 & 0.4920 & 0.4432 \\
GPT 4.1       & 0.4570 & 0.4380 & 0.4880 & 0.4730 \\
Claude Haiku  & 0.3967 & 0.3887 & 0.3890 & 0.4075 \\
Command R     & 0.3880 & 0.3700 & 0.4037 & 0.3840 \\
\bottomrule
\end{tabular}
\caption{Evaluation of choices-only accuracy on MMLU when adding ``I don't know'' (IDK) as one of the choices, similar to \citet{wang2025llms}. Accuracy still exceeds random guessing. \label{appendix:mmlu_idk}}
\end{table*}

\begin{table*}[h]
\centering
\small
\setlength{\tabcolsep}{3pt}
\begin{tabular}{lcccc}
\toprule
Model & Choices-Only (Ours) & Choices-Only (Inspect) & Choices-Only Reason (Ours) & Choices-Only Reason (Inspect) \\
\midrule
G-Lite        & 0.4530 & 0.4542 & 0.4840 & 0.4128 \\
G-Flash       & 0.5010 & 0.4930 & 0.5350 & 0.4908 \\
GPT-5 Mini     & 0.4640 & 0.4790 & 0.5290 & 0.4935 \\
GPT-4.1        & 0.4910 & 0.5090 & 0.5180 & 0.5175 \\
Claude-Haiku   & 0.4330 & 0.4258 & 0.4440 & 0.4198 \\
Command-R & 0.4530 & 0.4855 & 0.4440 & 0.4590 \\
\bottomrule
\end{tabular}
\caption{Comparison of the prompt used in this paper versus the standard MCQA prompt from InspectAI \citep{inspect_ai_framework}. We find little differences between these prompts, justifying our prompt design.}  \label{appendix:arc_prompts}
\end{table*}

\begin{table*}[h]
\centering
\small
\setlength{\tabcolsep}{3pt}
\begin{tabular}{lcccc}
\toprule
Model & Choices-Only (Ours) & Choices-Only (Inspect) & Choices-Only Reason (Ours) & Choices-Only Reason (Inspect) \\
\midrule
G-Lite        & 0.4170 & 0.4070 & 0.4440 & 0.3937 \\
G-Flash       & 0.4650 & 0.4447 & 0.4530 & 0.4193 \\
GPT-5 Mini     & 0.4258 & 0.4422 & 0.4920 & 0.4450 \\
GPT-4.1        & 0.4570 & 0.4765 & 0.4880 & 0.4532 \\
Claude Haiku   & 0.3967 & 0.4022 & 0.3882 & 0.3890 \\
Command-R & 0.3880 & 0.4085 & 0.4032 & 0.4037 \\
\bottomrule
\end{tabular}
\caption{Comparison of the prompt used in this paper versus the standard MCQA prompt from InspectAI \citep{inspect_ai_framework}. We find little differences between these prompts, justifying our prompt design.} 
\label{appendix:mmlu_prompts}
\end{table*}

\begin{trace}[title={Trace \thetcbcounter: Reasoning Trace Faithfulness Test --- Duplicated Choice}, label=trace:duplicate]

I need to analyze these multiple choice options without seeing the question. Let me look at what's provided:\\ \\

(A) a proton.\\(B) a neutron.\\(C) an electron.\\(D) an electron.\\ \\

I notice that options C and D are identical - both say "an electron." This is unusual for a multiple choice question, as typically each option would be different.\\ \\In most well-constructed multiple choice questions, having two identical options would be a formatting error. However, since I'm asked to select the correct answer, I need to think about what this could mean.\\ \\Possible interpretations:\\1. This is an error and C and D should be different\\2. The question is designed this way intentionally\\3. One of the electron options is the correct answer\\ \\Given that I have to choose an answer and both C and D are identical, if "an electron" is the correct answer to whatever question this is, then both C and D would technically be correct. However, since I need to pick one letter, and this seems like it could be a question about subatomic particles (given the options are proton, neutron, and electron), I should consider what's most likely.\\ \\Many physics questions involve electrons as answers - they're involved in electricity, chemical bonding, atomic structure, etc. The fact that "electron" appears twice might suggest it's the intended answer.\\ \\Between C and D, if I have to pick one, I'll go with C since it appears first chronologically.
\end{trace}

\begin{trace}[title={Trace \thetcbcounter: Reasoning Trace Faithfulness Test --- Synonymous Choice}, label=trace:synonym]

I need to analyze these choices to determine what the question might be asking about. Looking at the options, they all contain cellular organelles and structures:\\ \\(A) vacuole, chloroplast, nucleus\\(B) cell membrane, nucleus, mitochondrion\\(C) cell membrane, nucleus, and mitochondrion\\(D) mitochondrion, vacuole, cell wall\\ \\The key differences I notice:\\- Options B and C are nearly identical, with C just adding "and" before mitochondrion\\- Option A includes chloroplast, which is specific to plant cells\\- Option D includes cell wall, which is also specific to plant cells\\- Options B and C focus on structures found in all eukaryotic cells\\ \\This looks like it could be asking about:\\1. What organelles are found in all eukaryotic cells?\\2. What are the basic components of a cell?\\3. What structures are essential for cell function?\\ \\If it's asking about structures found in ALL eukaryotic cells (both plant and animal), then:\\- Vacuoles are much larger in plant cells, smaller in animal cells\\- Chloroplasts are only in plant cells\\- Cell walls are only in plant cells\\- Cell membrane, nucleus, and mitochondrion are found in all eukaryotic cells\\ \\Between options B and C, they're essentially the same content, but C has better grammar with "and" included in the list.\\ \\If this is about basic cellular structures that all eukaryotic cells have, then C would be the most complete and grammatically correct answer.
\end{trace}

\begin{trace}[title={Trace \thetcbcounter: Reasoning Trace Faithfulness Test --- Random Choice}, label=trace:random]
I need to analyze these multiple choice options to determine what the question might be asking and which answer is most likely correct.\\ \\Looking at the options:\\(A) to help scientists in conservation efforts\\(B) to determine the geographical habitats of all organisms  \\(C) banana\\(D) to understand the role of organisms in a food web\\ \\Options A, B, and D all seem to be related to scientific/biological purposes or goals. They all start with "to" followed by scientific objectives:\\- A: conservation efforts\\- B: determining geographical habitats\\- D: understanding organisms' roles in food webs\\ \\Option C "banana" is completely different - it's just a single noun referring to a fruit. This stands out as very different from the other three options.\\ \\In multiple choice questions, usually all the options relate to the same topic or theme. The fact that "banana" doesn't fit this pattern at all suggests this might be a trick question or the banana might be an obviously incorrect "distractor" answer.\\ \\However, given that three of the four answers follow a clear pattern of being scientific purposes/goals, and they're all reasonable scientific objectives, the question is likely asking about some scientific purpose or methodology. \\ \\Between A, B, and D:\\- A (conservation efforts) is a broad application\\- B (geographical habitats of ALL organisms) seems overly broad and ambitious\\- D (role of organisms in food web) is more specific and focused\\ \\D seems like the most reasonable and focused scientific objective among the legitimate options.
\end{trace}

\begin{trace}[title={Trace \thetcbcounter: Reasoning Trace Faithfulness Test --- Inaccurate Choice}, label=trace:inaccurate]

I need to analyze these choices about tectonic plates to determine which one is correct, even though I don't have the original question.\\ \\Let me evaluate each choice:\\ \\(A) "All tectonic plates come from asteroids" - This is clearly false. Tectonic plates are part of Earth's lithosphere and are not formed from asteroids. They are formed through geological processes on Earth.\\ \\(B) "Tectonic plates are in space" - This is also false. Tectonic plates are part of Earth's crust and upper mantle, located on Earth's surface, not in space.\\ \\(C) "Tectonic plates are composed of the oceanic lithosphere" - This is partially true but incomplete. Tectonic plates are composed of lithosphere, which includes both oceanic and continental lithosphere. Some plates are purely oceanic, some are purely continental, and some are mixed. So this statement is not entirely accurate as it suggests all plates are only oceanic lithosphere.\\ \\(D) "Tectonic plates are not linked to earthquakes" - This is false. The movement and interaction of tectonic plates at their boundaries is the primary cause of earthquakes. This is a fundamental principle in geology and seismology.\\ \\Looking at these options, (C) seems to be the closest to being correct, even though it's not perfectly stated. Tectonic plates are indeed composed of lithosphere, though not exclusively oceanic lithosphere. All the other options are clearly false statements about basic geological facts.\\ \\However, let me reconsider (C) more carefully. The statement says tectonic plates are composed of "the oceanic lithosphere." This could be interpreted as saying they are made of oceanic lithosphere material, but it's not saying ALL plates are ONLY oceanic. It could be a general statement about the composition being lithospheric material, with oceanic being used as an example or primary type.\\ \\Still, this seems like the most reasonable answer among clearly incorrect alternatives.
\end{trace}
\hypersetup{
    linkcolor=white,
    citecolor=white,
    urlcolor=white
}

\lstset{
  literate={<}{{<}}1
           {>}{{>}}1
}

% --- Spanning prompt 1 (kept text the same; just use the env directly) ---

\begin{prompt}[title={Prompt \thetcbcounter: Full MCQA Prompt}, label=prompt:full]
You will be given a question and a set of choices from a multiple-choice question. Your goal is to generate reasoning to select the letter of the correct answer for the question. Use any strategy necessary to come up with the correct answer

\{input\_text\}

Generate step-by-step reasoning to perform this task. Your final response should be in the format:
<reasoning>
[insert reasoning]
</reasoning>
<answer letter>
[insert letter of the correct answer]
</answer letter>
\end{prompt}

\begin{prompt}[title={Prompt \thetcbcounter: Partial MCQA Prompt}, label=prompt:partial]
You will be given just a set of choices from a multiple-choice question. Your goal is to generate reasoning to select the letter of the correct answer, even though you do not have access to the question. Use any strategy necessary to come up with the correct answer

\{input\_text\}

Generate step-by-step reasoning to perform this task. Your final response should be in the format:
<reasoning>
[insert reasoning]
</reasoning>
<answer letter>
[insert letter of the correct answer]
</answer letter>
\end{prompt}

\begin{prompt}[title={Prompt \thetcbcounter: Full MCQA Prompt with Chain-of-Thought}, label=prompt:full_cot]

You will be given a question and a set of choices from a multiple-choice question. Your goal is to generate reasoning to select the letter of the correct answer for the question. Use any strategy necessary to come up with the correct answer

\{input\_text\}

Generate step-by-step reasoning to perform this task. Your final response should be in the format:
<reasoning>
[insert reasoning]
</reasoning>
<answer letter>
[insert letter of the correct answer]
</answer letter>
\end{prompt}

\begin{prompt}[title={Prompt \thetcbcounter: Choices-Only MCQA Prompt with Chain-of-Thought}, label=prompt:partial_cot]
You will be given just a set of choices from a multiple-choice question. Your goal is to generate reasoning to select the letter of the correct answer, even though you do not have access to the question. Use any strategy necessary to come up with the correct answer

\{input\_text\}

Generate step-by-step reasoning to perform this task. Your final response should be in the format:
<reasoning>
[insert reasoning]
</reasoning>
<answer letter>
[insert letter of the correct answer]
</answer letter>
\end{prompt}

\begin{trace}[title={Trace \thetcbcounter: Reasoning Trace --- Inferring the Original Question}, label=trace:itq]

**Deconstructing the Multiple Choice Puzzle: My Analysis**\\

Alright, let's break this down. First, I'm staring at a multiple-choice problem with no context. My job is to pick the correct letter using whatever strategies I have. The format is locked – I have to give the answer as `<answer letter>[letter]`.  I'm going to approach this with a few angles.\\

Initially, I'm observing the provided options: (A) oil, (B) coal, (C) trees, and (D) aluminum. A good starting point is categorization. Looking at these, I immediately see patterns: Oil and coal are both fossil fuels; trees are a biological resource, and aluminum is a metal. I can categorize them further like this: oil, coal, and trees are all organic materials, with aluminum as the odd inorganic one. Also, three are non-renewable (on human timescales), and one, trees, is renewable. They're all *sourced* resources, but from different origins.\\

Now, let's explore the "odd one out" strategy.  Here, the key is looking for one option that stands apart.\\

*   **Renewable vs. Non-renewable:** Trees stick out as the lone renewable option. The others are non-renewable resources we're actively depleting. This is a solid contender.
*   **Organic vs. Inorganic:** Aluminum, being a metal, is inorganic. The others have an organic origin.
*   **Fossil Fuel vs. Other:**  Oil and coal are both fossil fuels. Trees and aluminum aren't.
*   **Source:** I am also considering where each comes from.  Oil and coal come from ancient organic material, trees are living plants, and aluminum comes from ore.\\

So, the next question becomes: what are questions likely to be asked about these?  The question "Which of the following is a renewable resource?" immediately springs to mind, making C (trees) a very probable answer.  A metal question "Which of the following is metal?", making D a strong contender, is plausible.  For A and B (oil and coal), I am looking for a question to help me distinguish.\\

I'm now weighing the "odd one out" possibilities. C (trees) with the renewability distinction, and D (aluminum), the metal, are strong choices. My internal discussion leans towards C (trees) because the renewable/non-renewable distinction feels a bit more foundational and prevalent in general knowledge and education.  I often see these grouped as "finite resources." The organic/inorganic distinction is also valid, but slightly less common in these sets of resources.\\

Another check: what comes to mind when I see each term? "Trees" immediately bring to mind "renewable" and "paper," making "trees" and C very strong.  "Aluminum" is associated with "recyclable", "metal" etc.  Both look strong.\\

In summary, the distinction is clearest with renewability. The question, "Which of the following is a renewable resource?" is a common and straightforward educational problem. The grouping of oil, coal, and aluminum is often grouped as non-renewable resources, and trees naturally fit. While the question could be inorganic/metal. The first has a higher probability.\\

Ultimately, between the two strong contenders, my professional instincts lean towards C, because of the likelihood of using it for a question about renewability.\\

Therefore, my best guess is: **<answer letter>C</answer letter>**
\end{trace}

\begin{trace}[title={Trace \thetcbcounter: Reasoning Trace --- Finding Dataset Flaws}, label=trace:flaw]

**Initial Assessment and Strategy Development**\\

Okay, here's what I'm dealing with: I've got a multiple-choice question with four statistical visualization options, but *no* question. The instructions are clear - I need to choose the "correct" letter based on *any* strategy, implying that I'm supposed to employ a bit of logical deduction or pattern recognition rather than just randomly guessing. The output needs to be in a specified XML format.\\

My first thought is to break down the choices and look for patterns. I'm seeing: (A) bar graph, (B) line graph, (C) scatterplot, and (D) stem and leaf. All four are common methods of displaying data, but I need to differentiate them somehow.\\

**Analyzing the Options: Categorization and Differentiation**\\

I'll start by categorizing them:\\

*   (A) Bar graph: Best for comparing categorical data.
*   (B) Line graph: Ideal for visualizing trends over time or continuous data.
*   (C) Scatterplot: Shows the relationship between two numerical variables.
*   (D) Stem and leaf: Displays quantitative data while preserving the original values, essentially a visual representation of a data distribution.\\

My initial approach is to see if I can find an outlier. A, B, and C are all common and versatile graph types. A stem-and-leaf plot is a more specific and less frequently encountered visual tool. It's essentially a tabular and graphical hybrid, making it distinct from the pure graphical representations. This could potentially make (D) a stronger candidate.\\

**Strategic Considerations: "Guessing" without a Question**\\

Since I lack the question, my primary strategy has to be the most likely test-taking situation: The test designer wants to differentiate between these chart types. So, let's look at the "meta" game:\\

*   **Commonality/Outlier Strategy:** Bar graphs, line graphs, and scatterplots are very common. A stem-and-leaf plot is less so, making it stand out.
*   **Frequency/Typicality:** I need to consider which are commonly confused, like when to use a line graph vs. scatterplot. However, without a question context, this is a dead-end.\\

**Developing Hypotheses and Reframing**\\

Now, I'll try framing the situation from different perspectives. I think the key is understanding the type of data or purpose.\\

*   A, B, and C represent ways to visualize relationships between variables.
*   D focuses on the distribution of a single variable. This difference seems significant. Maybe the question is about *which visualization method best shows a dataset's distribution.*\\

**Revisiting "Meta" and Eliminating Other Options**

The idea here is that there could be a "trick" question. It could involve commonly confused terms, and the key might be finding the option that's the most distinct.\\

Looking at each one:\\

*   **A: Bar Graph:** Best for comparing categories
*   **B: Line Graph:** Best for showing a trend in continuous data, such as time
*   **C: Scatterplot:** Best for showing relationships between variables and their correlation
*   **D: Stem-and-Leaf:** Best for displaying a single numerical data set and its distribution\\

Based on the nature of "test" and "guessing without a question," I am left with a choice: I think "Stem and Leaf" stands out, as it is a less common visualization method. Therefore, this is the most likely choice for a "correct" answer in the absence of additional information or an actual question.
\end{trace}

\begin{trace}[title={Trace \thetcbcounter: Reasoning Trace --- Finding Patterns in Choices}, label=trace:patterns]

The choices provided are numerical values: -1, +1, -2, and +2. These are all small integers, symmetrically distributed around zero, suggesting a context involving signs, direction, parity, or possibly eigenvalues, charges, or mathematical operations such as determinants or exponents.\\

Notably, +1 and -1 are common in contexts involving:
- Multiplicative identities
- Roots of unity
- Parity (odd/even functions or permutations)
- Dot products or normalization
- Direction (left/right, positive/negative feedback)\\

+2 and -2 are less common in fundamental identities and typically arise as sums, differences, or specific values in equations.\\

In many multiple-choice contexts—especially in mathematics, physics, or standardized tests—when choices include symmetric values around zero, the correct answer is often +1 or -1, particularly if the question involves normalization, identity properties, or sign determination.\\

Moreover, +1 is a frequent answer in questions involving:
- The value of expressions like cos(0), e**(i0), or determinant of identity matrix
- Net change that cancels out to a unit value
- Multiplicative inverses
- Positive outcomes in balance or symmetry\\

Given the absence of the question, we consider frequency heuristics: +1 appears more commonly than -1 as a correct answer in neutral contexts, and +2 or -2 are typically associated with more specific scenarios (e.g., charge of alpha particle = +2), but without context, such specificity is less likely.\\

Additionally, if this were a question about the sign of a quantity that is known to be positive (e.g., magnitude, squared term, absolute value), +1 or +2 would be candidates. But +1 is more canonical.\\

Another angle: if the question were about the difference in degrees of freedom, or a net gain/loss, +1 is a common minimal positive increment.\\

Considering symmetry, simplicity, and frequency in test design, +1 stands out as the most likely correct answer among the options.\\

Thus, based on reasoning from typical patterns in multiple-choice questions, the most plausible correct answer is (B) +1.
\end{trace}

\end{document}